\documentclass[final,12pt]{clear2024} 

\usepackage{mathtools}
\usepackage{mathabx}
\usepackage[utf8]{inputenc} 
\usepackage[T1]{fontenc}    
\usepackage{hyperref}       
\usepackage{booktabs}
\usepackage{tabulary}       
\usepackage{amsfonts}       
\usepackage{bbm}
\usepackage{nicefrac}       
\usepackage{microtype}      
\usepackage{color}
\usepackage{algorithm}
\usepackage{algorithmicx}
\usepackage{algpseudocode}
\usepackage{tikz}
\usepackage{kbordermatrix}
\usepackage{longtable}

\usepackage{pgfplots}
\pgfplotsset{compat=1.16}
\usepackage{wrapfig}
\usepackage{array}
\usetikzlibrary{fit,positioning,calc,arrows,backgrounds}
\newcolumntype{C}[1]{>{\centering\arraybackslash}p{#1}}
\tikzset{
    main/.style = {
        draw,
        circle,},
    every node/.style={scale=.7}
}

\usepackage{tkz-berge}
\usepackage{tikz-3dplot}

\usetikzlibrary{patterns}
\usetikzlibrary{patterns}
\usetikzlibrary{decorations.pathreplacing}
\usetikzlibrary{graphs,graphs.standard}
\usetikzlibrary{shapes.geometric,backgrounds}
\usetikzlibrary{patterns.meta}
\usetikzlibrary{matrix}
\usetikzlibrary{fadings}
\usetikzlibrary{trees}
\usetikzlibrary{mindmap}
\usetikzlibrary{arrows.meta}
\GraphInit[vstyle=Classic]
\SetGraphUnit{1}
\SetVertexMath

\definecolor{applegreen}{rgb}{0.0,0.5,0.0}
\definecolor{spgreen}{RGB}{46,139,87}
\definecolor{tored}{RGB}{220,20,60}
\newcommand{\cupdot}{\mathbin{\mathaccent\cdot\cup}}

\usepackage{titleps}
\newpagestyle{group}{%
\headrule
\sethead{}{}{}%
\setfoot{}{\thepage}{}}

\usepackage[singlelinecheck=off]{caption}

\title[\texttt{causalAssembly}]{\texttt{causalAssembly}: Generating Realistic Production Data for Benchmarking Causal Discovery}
\usepackage{times}

\clearauthor{%
 \Name{Konstantin G{\"o}bler} \Email{konstantin.goebler@tum.de}\\
 \addr Technical University of Munich, Germany \\
 Robert Bosch GmbH%
 \AND
 \Name{Tobias Windisch} \Email{tobias.windisch@hs-kempten.de} \\
 \addr University of Applied Sciences Kempten%
 \AND
 \Name{Mathias Drton} \Email{mathias.drton@tum.de}\\
 \addr Technical University of Munich, Germany \\ Munich Center for Machine Learning (MCML)%
 \AND
 \Name{Tim Pychynski} \Email{tim.pychynski@de.bosch.com}\\
 \Name{Steffen Sonntag} \Email{steffen.sonntag@de.bosch.com}\\
 \Name{Martin Roth} \Email{martin.roth2@de.bosch.com}\\
 \addr Robert Bosch GmbH
}

\begin{document}

\maketitle
\thispagestyle{group}

\begin{abstract}%
    Algorithms for causal discovery have recently undergone rapid advances and increasingly draw on flexible nonparametric methods to process complex data. With these advances comes a need for adequate empirical validation of the causal relationships learned by different algorithms. However, for most real and complex data sources true causal relations remain unknown. This issue is further compounded by privacy concerns surrounding the release of suitable high-quality data. To tackle these challenges, we introduce \texttt{causalAssembly}, a semisynthetic data generator designed to facilitate the benchmarking of causal discovery methods. The tool is built using a complex real-world dataset comprised of measurements collected along an assembly line in a manufacturing setting. For these measurements, we establish a partial set of ground truth causal relationships through a detailed study of the physics underlying the processes carried out in the assembly line. The partial ground truth is sufficiently informative to allow for estimation of a full causal graph by mere nonparametric regression. To overcome potential confounding and privacy concerns, we use distributional random forests
to
estimate and represent conditional distributions implied by the ground truth causal graph. These conditionals are combined into a joint distribution that strictly adheres to a causal model over the observed variables. Sampling from this distribution, 
\texttt{causalAssembly} generates data that are guaranteed to be Markovian with respect to the ground truth. Using our tool, 
we showcase how to benchmark several well-known causal discovery algorithms.

\end{abstract}



\section{Introduction}\label{intro}

Causal discovery is the process of learning causal relations among variables of interest from data. In recent years, the area has seen numerous theoretical and algorithmic developments \citep[for an overview see, e.g.][]{Glymour2019,Heinze18,Maathuis2019,Vowels2022, Zanga2022}. Causal structure learning is also gaining traction in industrial domains \citep[compare e.g.~the review in][]{Vukovic2022}, where knowledge about causal relations forms the basis for downstream operations such as root-cause analysis \citep{Budhathoki2022}.

Despite the recent advances and expanded application areas, it remains challenging to provide adequate empirical validation of the causal relationships inferred by causal discovery algorithms \citep{Gentzel2019,Eigenmann2020}. Ground truth causal relationships are unknown for most real-world examples that involve complex data sources. In addition, privacy concerns often hinder the release of high-quality data. As a result, research papers proposing novel procedures often
work with simplistic simulation setups that
tend to have
limited generalizability to real-world settings \citep{Huegle2021,Reisach2021}. When applied to real data and assessed by domain experts, results are often sobering \citep{Heinze18,Constantinou2021}.

To help address these challenges we introduce \texttt{causalAssembly}, a semisynthetic data generation tool that leverages extensive domain knowledge and real production data to form a ground truth. The assembly line that originates the data is composed of multiple production stations where individual components are joined together through automated manufacturing processes. Each individual manufacturing process highly depends on the state of the preceding processes as well as on the raw components added in earlier stations. Due to the intricate and often nonlinear nature of the physical processes---including, e.g., the interaction of machine state, process control settings and component state, complex causal relationships arise across the entire assembly line. In the manufacturing plant, processes are computer-monitored, and the resulting measurements are stored throughout production.



We select a subset of production stations from the assembly line, each of which contains individual processes with known causal relationships. We determine these relationships by consulting with domain experts and carefully studying the physical processes that take place at each station. However, our process knowledge currently does not extend to relationships between processes. We show that the extensive domain knowledge of the individual processes, together with the line structure, gives rise to a causal ordering that agrees with that of the unknown ground truth layered graph. Consequently, we apply established feature selection strategies \citep[e.g.][]{Buhlmann2014} to prune the corresponding complete graph implied by this causal ordering. The resulting causal graph is treated as the appropriate ground truth in \texttt{causalAssembly}.

Next, we need to enable sampling from a ground truth distribution that factorizes according to the ground truth causal graph. This is achieved with a synthetization step that learns conditional distributions indicated by the ground truth graph via distributional random forests (DRF) \citep{Cevid2022,Gamella2022}. Combining the conditionals yields a joint distribution that strictly obeys a causal model for the observed variables. Thus, the  distribution of semisynthetic data sampled with \texttt{causalAssembly} is guaranteed to be Markov with respect to the ground truth graph. 
We want to note that strong-faithfulness \citep[see][]{Uhler2013} can not be guaranteed to hold in the sampled data which may affect consistency results of some causal discovery algorithms.

\smallskip 
\textbf{Our contributions.} We summarize our main contributions below:
\begin{itemize}
    \item Using domain knowledge and real production data, we propose a semisynthetic data generation pipeline \texttt{causalAssembly}. Implemented as a Python library\footnote{\url{https://github.com/boschresearch/causalAssembly}}, the tool allows one to generate data of varying dimension and complexity in order to benchmark causal discovery algorithms.
    \item The proposed procedure to synthesize real data is general and can in principle be applied to any real dataset with full or partial domain knowledge. The data generation procedure assures resulting data to be Markovian under causal sufficiency.
    \item The proposed semisynthetization of real data offers a way to release high-quality data to the public that would otherwise be inaccessible due to privacy concerns. To employ this procedure, knowledge regarding a causal order will be 
          a necessary requirement.
    \item We provide first benchmark results of popular causal discovery algorithms including the PC algorithm \citep{Spirtes1991}, DirectLiNGAM \citep{Shimizu2011}, NOTEARS \citep{Zheng2018}, GraN-DAG \citep{Lachapelle2020}, and SCORE (and its scalable extension DAS) \citep{Rolland2022,Montagna2023}.
\end{itemize}
In the remainder of the paper, we first address identifiability concerns and provide an overview of related work (Section~\ref{sec:prelim}). Section~\ref{sec:datagen} describes the data generation pipeline, and
Section~\ref{sec:benchmark} showcases benchmarking results. Section~\ref{sec:conclusion} gives a concluding discussion of our work. Supplementary Sections~\ref{sec:app_a}-\ref{sec:app_c} provide additional information regarding data, algorithms and metrics used, as well as additional benchmark results.


\section{Identifiability and common benchmarking strategies}\label{sec:prelim}

\subsection{General background}\label{sec:general}


For natural number
$p\in\mathbb{N}$, let $[p]:=\{1,\ldots, p\}$. Defining the vertex set as \(V = [p]\), let
\(G=(V,E)\) be a directed
acyclic graph (DAG) with edge set \(E \subset V \times V\).
The vertex set 
\(V\) indexes a
collection of jointly distributed random variables \((X_v)_{v\in V}\). For 
$S\subseteq V$, we write $X_S:=(X_s)_{s\in S}$ for the subcollection of
variables indexed by $S$.
We often write \(v
\rightarrow w\) or $X_v\to X_w$  instead of \((v,w)\) to indicate the existence
of an edge from $v$ to $w$ in $G$. A node \(v\in V\) is a \textit{parent} of
another node \(w\in V\) if \(v \rightarrow w \in E\). The set of all
parents of \(w\) in \(G\) is denoted by \(pa_G(w)\). A \textit{directed path}  from $v$
to $w$ in $G$ is a sequence of nodes $v_1,\dots,v_k$ starting at $v_1=v$ and ending at $v_k=w$
with subsequent nodes connected by edges $v_i\to v_{i+1}\in E$.
If there exists a directed path from \(v\) to \(w\), then \(v\) is an \textit{ancestor} of \(w\) and $w$ a
\textit{descendant} of $v$.
The sets of all ancestors and all descendants of node \(w\) in \(G\) are
denoted \(an_G(w)\) and \(de_G(w)\), respectively and \(w \notin an_G(w) \cup de_G(w)\). When it is  clear which graph $G$ is being considered, we simply write $pa(w)$, $de(w)$ or
$an(w)$.

The DAG \(G\) permits causal relationships among the random
variables, and we say that \(X_v\) is a direct cause of \(X_w\) if \(v \rightarrow
w\). Furthermore, \(X_v\) may be an indirect cause of \(X_w\) if there
is a directed path from \(v\) to \(w\); we refer
to~\cite{Pearl2009,Peters2017} for more background. Each DAG \(G\) on $V$ gives rise to a statistical model for the joint distribution \(P_X\) of the random vector \(X=(X_v)_{v \in V}\).  Here, the edges in
\(G\) encode the permitted conditional dependence relations among the
variables \citep[see e.g.][]{Drton2017,Lauritzen1996}. Any distribution
that satisfies the conditional independence constraints implied by
\(G\) is said to be \textit{Markov} with respect to \(G\). In
other words, the joint density of \(X\) factors into a product of
conditional densities as
\(
p(x) =
\prod_{v \in V} p(x_v \mid x_{pa_G(v)})
\)
with \(p(x_v\mid
x_\emptyset) = p(x_v)\).
The conditional independence relations holding in all such factorizing distributions can be obtained using the graphical criterion of \textit{d-separation}
\citep{Pearl2009}. A distribution \(P_X\) is \textit{Markov} and
\textit{faithful} with respect to the underlying DAG \(G\) if the conditional
independence relations in \(P_X\) correspond exactly to
d-separation relations in $G$.
Different DAGs may share the same set of
conditional independence statements. Such DAGs are
said to be Markov equivalent. All equivalent DAGs form a Markov
equivalence class (MEC) \citep[see e.g.][]{Spirtes1993}.

\subsection{Layered DAGs}

Let $G=(V, E)$ be a DAG, and let \(\mathcal{L} := (V_1, \dots, V_K)
\) be a partition of \(V\) into \(K\) ordered subsets. Then \(L =
(G,\mathcal{L})\) is a \emph{layered DAG} if
for every edge \((v,w) \in E\) with nodes \(v \in V_s\), \(w \in V_t\)
with $s\neq t$,
we have $s<t$ \citep[see, e.g.,][]{Healy2002}.  In this
case, the elements of $\mathcal{L}$ are the \emph{layers} of $L$, and each layer $V_s$ induces a subgraph \(G[V_s]\) of $G$, which we denote by \(L_s\) and refer to as the layer-induced subgraph.

A permutation \(\pi: [p] \to [p]\) is a causal ordering for the layered DAG \(L =(G,\mathcal{L})\) if (i) \(\pi(v) < \pi(w)\) for all $(v,w)$ with \(w \in de_G(v)\), and (ii) \(\pi(v) < \pi(w)\) for all $(v,w)$ such that $v\in V_s$ and $w\in V_t$ for layers with $s<t$.
Such an ordering respects both the ordering given by directed paths in $G$ and the ordering of the layers in $\mathcal{L}$.  There always exist a causal ordering but it need not be unique.
The set of all causal orderings of a layered DAG $L$ is denoted by $\Pi_L$.

\begin{proposition}\label{prop:same_causal_orders}
    Let \(L = (G,\mathcal{L})\) and \(L^{\prime} = (G',\mathcal{L})\) be two layered DAGs that share the same vertex set $V$ and the same partition $\mathcal{L}=(V_1,\dots,V_K)$.  If \(L_s = L^\prime_s\) for all \(s \in [K]\), then \(\Pi_L = \Pi_{L^{\prime}}\).
\end{proposition}
The proof can be found in Section \ref{sec:proofs}. Let $L=(G,\mathcal{L})$ be a layered DAG with vertex set $V$ and edge set $E$.
For any causal ordering \(\pi \in \Pi_L\), we define a completed layered DAG \(L^\pi =(G^\pi, \mathcal{L})\) by adding to $L$ all edges
$v\to w$ with $v\in V_s$ and $v\in V_t$ for $s<t$.  So, the edge set of $L^\pi$ is
\[
    E^\pi := E\cup \bigcup_{s<t} V_s\times V_t.
\]
Clearly, $L^\pi$ is a super-DAG of $L$ in the sense of $E$ being a subset of $E^\pi$.
Often we have to consider all nodes up to a certain layer. Thus, we define $V_{1:s}:=\cup_{i=1}^{s-1}V_i$ for $s\in[K]$. Note that $V_s\cap V_{1:s}=\emptyset$.


\subsection{Identifiability of the causal structure}\label{sec:identifiability}

In general, the DAG \(G\) is not identifiable from \(P_X\) alone, and learning the graph or its MEC from data
requires assumptions. Under the Markov and
faithfulness conditions on \(P_X\), \(G\) can be
recovered up to its MEC.
Prominent examples of algorithms that are based on this assumption are the PC algorithm \citep{Spirtes1993} and Greedy Equivalence Search (GES)
\citep{Chickering2003}.
Under additional assumptions, we may not only recover the MEC
but \(G\) itself. Note that the graphical model associated
with \(G\) can also be expressed in terms of a structural causal
model (SCM) \citep[][]{Peters2017}. That is, in the absence of latent confounding, if \(P_X\) is Markov with
respect to \(G\) then there exist independent noise variables
\(\epsilon = (\epsilon_1,\ldots, \epsilon_p)\) and measurable functions
\(g_{v}\) such that \(X_v = g_{v}(X_{pa(v)}, \epsilon_{v})\) for
all \(v \in [p]\).
In this framework, if all the \(g_{v}\) are linear and all \(\epsilon_{v}\) follow
non-Gaussian distributions
then the causal graph is identifiable from data
\citep{Shimizu2022}.
In a different vein, identifiability of $G$ is also achieved under an additive noise model
(ANM) \citep{Hoyer2008}. In other words, as long as all
\(g_{v}\) are non-linear as defined in~\citet{Hoyer2008}, the noise \(\epsilon_{v}\) may follow
any distribution given that it is additive. A generalization to
the post-nonlinear model is proposed by \citet{Zhang2009}.
In contrast, linear SCMs with Gaussian noise terms are not
identifiable. However,
identifiability can be achieved in this setting under further assumptions such as equal noise variances \citep{Peters2014b}.

\subsection{Additive noise models for synthetic data generation}\label{sec:anms}

Comprehensive assessment of methods for learning causal relations is challenging
due to the scarcity of real data where true causal relations are
known.  Hence, the focus often lies on simulated synthetic data. In order to accommodate non-linearities and also to ensure identifiability of the true DAG, ANMs are often employed for this purpose.
However, synthetic data generation can leave unintended traces hinting at the true
causal relations among the variables. In particular,
\citet{Reisach2021} demonstrate that for commonly used simulation setups involving ANMs, irrespective of the choice of the functions \(g_{v}\), ordering nodes by
increasing marginal variance tends to agree with the causal ordering of the
ANM. The authors introduce \textit{varsortability}, a
score that measures the degree of concordance between causal orderings of
the ground truth DAG and the order of increasing marginal variance.
When varsortability equals one
then all directed paths lead from lower variance nodes to higher
variance nodes.
This is connected to the aforementioned identifiability results for linear SCMs with equal variance \citep{Chen2019}.
Many 
recently proposed continuous optimization-based structure
learning algorithms have been evaluated in simulations
based on ANMs \citep[e.g.,][]{Zheng2018, Ng2020, Lachapelle2020, Yu2021, Zhang2022, Zhang2023}. The resulting high varsortability on the raw data scale contributes to often staggering performances as the methods may inadvertently make use of variance order. In real systems, on the
other hand, measurement scales are essentially irrelevant for the
causal ordering. Thus, standardization would 
render data scale non-informative. Yet, remnants of high
varsortability with distinctive covariance patterns
may remain even after standardization~\citep{Reisach2021,Reisach2023}.
Ultimately, all synthetic data generation processes are
vulnerable to a lack of portraying common themes observable in
practice.

Evaluating the performance of algorithms
is a key factor that impacts which research directions the field is
gravitating towards. This can be seen, for instance, in the case of
continuous optimization-based structure learning. Following the
development of NOTEARS \citep{Zheng2018}, numerous extensions have
been proposed in short succession \citep[see][for an
    overview]{Vowels2022}. However, when evaluated outside of ANMs, performance often drops notably \citep{Constantinou2021}.
Therefore, there is a clear necessity for more robust evaluation
techniques in the field of causal discovery as outlined
in~\citet{Eigenmann2020,Gentzel2019}.
This motivates the present work which seeks to support benchmarking of causal discovery by providing tools for semisynthetic data generation that draw on partial ground truth and real data from a manufacturing plant.

\subsection{Related work}

Focusing on observational data, the available tools for empirical illustration and benchmarking of causal discovery algorithm can be broadly classified into three tiers:
(1) Real datasets with expert-knowledge driven
ground truth, such as the cause-effect pairs of \cite{Mooij2016} and data pertaining to 
already well understood systems in biology~\citep{Marbach2012, Replogle2022,Sachs2005}. (2)
Synthetic datasets generated based on well-established
parameterizations
\citep[][]{VandenBulcke2006, Schaffter2021, Scutari2010}. (3) Data
generated using functional specifications; see, e.g., \cite{Chen2023,Huegle2021}. Our proposed semisynthetic data generator takes a middle ground between (1) and (2), leveraging expert knowledge to establish a causal order while relying on real data to circumvent the need for explicit parameterizations.

\section{Semisynthetic data generator}\label{sec:datagen}

\subsection{Process knowledge}\label{sec:expertknowledge}

\begin{figure}
    \begin{minipage}{0.45\textwidth}
        \centering
        \includegraphics[width=.9\linewidth]{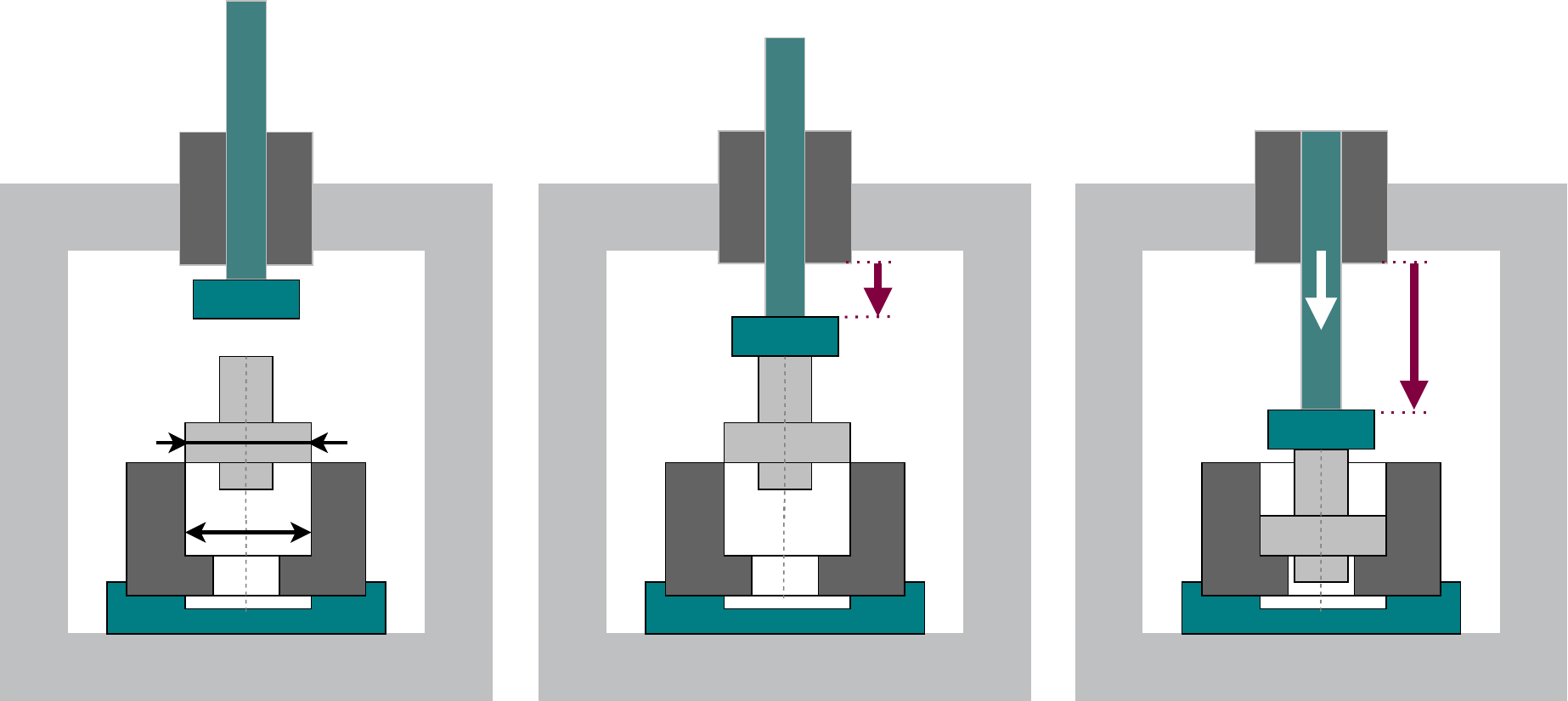}\\
        \textit{(a) Press-in}
    \end{minipage}
    \hfill
    \begin{minipage}{0.45\textwidth}%
        \centering
        \includegraphics[width=.9\linewidth]{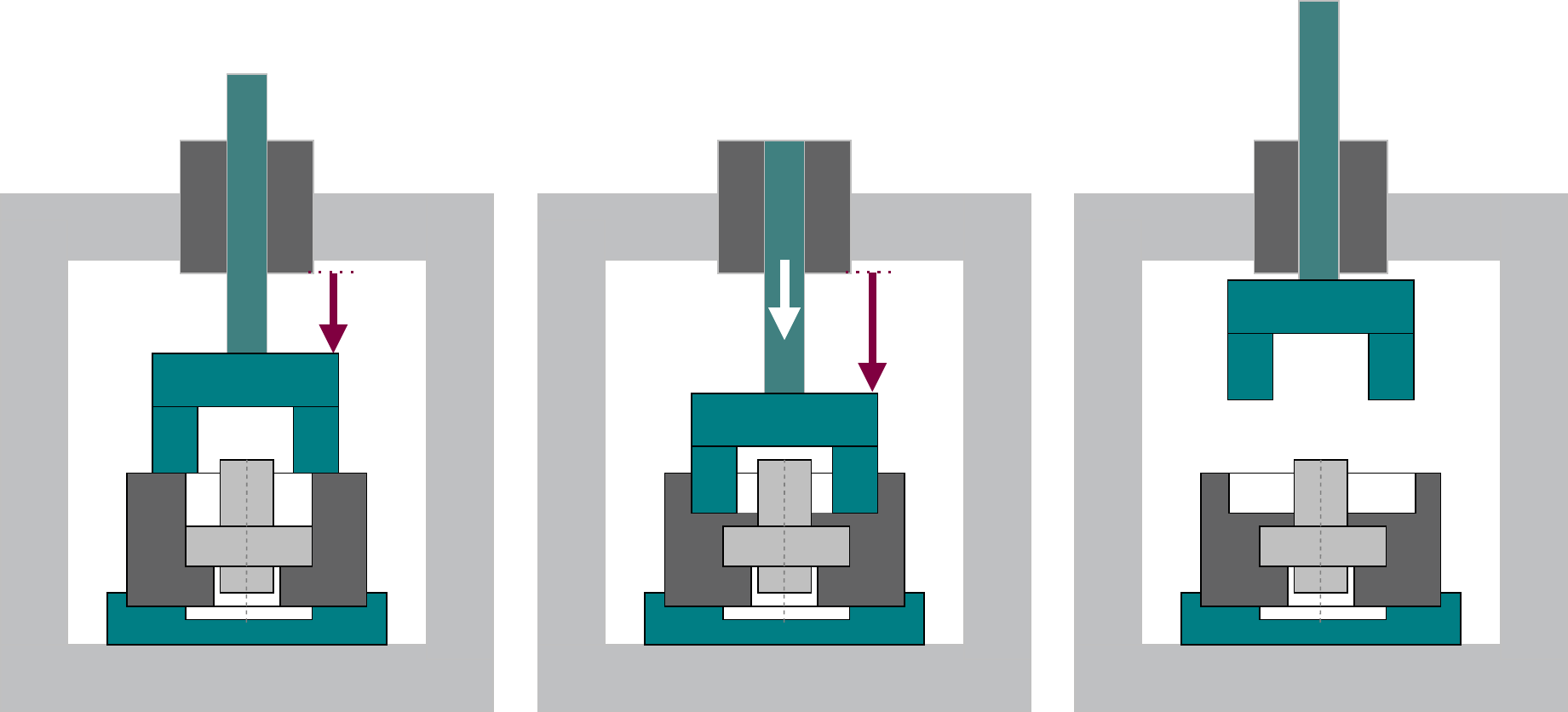}\\
        \textit{(b) Staking}
    \end{minipage}
    \caption{Illustration of the phases of a press-in and staking process. In press-fitting 
        (a1) the tool moves downward axially until it contacts the valve (a2) and pushes it into a slightly smaller bore until its shoulder reaches the axial block position (a3).
    }
    \label{fig:pressin_and_staking}
\end{figure}

Our real-world data come from an assembly line at a highly automated manufacturing plant.
We focus on a selected subset of production stations at which press-in and staking processes are executed. In the following, we give a brief overview of these
processes.  A more in depth discussion of the relations among the central
features of the physical procedures is given in supplementary Section~\ref{sec:process_knowledge}.

\textit{Press-in: } In each fitting process, two components are
joined in a pressing machine to form a mechanically strong and
sometimes even leakage tight connection between the components.  An
inner cylindrical component (e.g., a valve, axle or pin) is pushed into
a slightly smaller bore of a larger component. This is done by
applying a mechanical force high enough to overcome the required
friction force. After the friction force is exceeded, the inner
component will move into the bore. The pressing machine stops as soon
as a certain predefined press-in force is exceeded for the first time.

\textit{Staking: } A staking process is an extension of a press-in process and has the objective to fixate the inner component in its final position by permanently deforming outer component materials. This is achieved by applying a large enough force to the tool to create a mechanical stress-strain state in the material resulting in permanent plastic material deformation.

Figure~\ref{fig:pressin_and_staking} depicts these two process steps in a
schematic manner. During both steps the required forces and tool displacements are measured continuously by process control units. We extract all relevant numeric measurements for each process, and together with the mapping detailed in Section~\ref{sec:process_knowledge} they build the foundation for our data generation scheme.


\subsection{Modeling domain knowledge}

\begin{figure} 
    \centering

    \begin{tikzpicture}[scale=0.7]

        \foreach \i/\l in {1, 2, 3, 4, 5}{
                \draw[rounded corners, fill=gray!15] (4*\i-4-0.15, -0.25) rectangle (4*\i-.5+0.15, 2.50) {};
                \node[scale=1.2] at (4*\i-4+1.75, 2.2) {Station\i};
            }

        \foreach \i in {1, 2, 3, ..., 10}{
                \draw[rounded corners, fill=gray!55] (2*\i-2-0.05, 0) rectangle (2*\i-0.5+0.05, 1.5) {};
                \node[scale=1.0, align=center] at (2*\i-2+0.75, 1.15) {Process\i};
                \node[scale=1.4, align=center] at (2*\i-2+0.75, 0.45) {$V_{\i}$};
            };

        \draw[->, line width=1.5pt] ( 6, -0.7) to node[midway, fill=white] {component flow} (14, -0.7);

    \end{tikzpicture}

    \caption{Illustration of the production line with five process
        stations each containing two successive processes. The first station
        prepares components while station two and four carry out staking tasks. The remaining stations perform press-in tasks.}
    \label{fig:pline_schematic}
\end{figure}
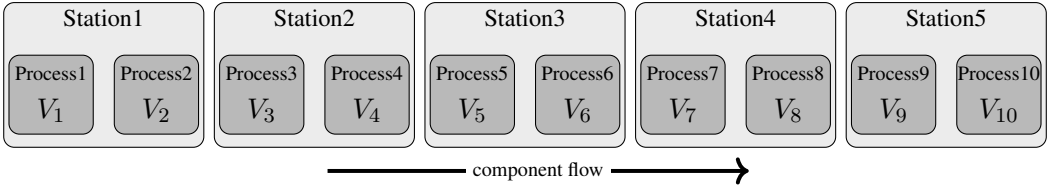

Let \(L^0 = (V, E^0, \mathcal{L})\) be the true and unknown layered
DAG that encapsulates within as well as between process relations. The
associated random variables \((X_v)_{v\in V}\) represent the features
extracted from the \(K=10\) manufacturing processes that are carried
out sequentially. These processes are lined up across five production
stations. A user of \texttt{causalAssembly} will see which variables belong to each of these five
stations.  Each station accommodates two of the production processes
(Fig.~\ref{fig:pline_schematic}). In total, \(p=\lvert V \rvert = 98\) features for \(n = 15.581\) parts are collected. The features distribute across production stations as follows: \textit{Station 1} with \(\lvert V_1 \cup V_2 \rvert = 6\), \textit{Station 2} with \(\lvert V_3 \cup V_4 \rvert = 34\), \textit{Station 3} with\(\lvert V_5 \cup V_6 \rvert = 16\), \textit{Station 4} with \(\lvert V_7 \cup V_8 \rvert = 26\), and \textit{Station 5} with \(\lvert V_9 \cup V_{10} \rvert = 16\).

Each individual process we focus on is well understood, and we are
able to state causal relations among the variables in $V_s$. These
causal relations are considered to be complete and correct for the
individual processes. We map the process knowledge into process DAGs
$G_s^*=(V_s, E_s)$ for $s\in[K]$. The graph union \(
G_{1}^{*} \oplus \dots \oplus G_{K}^*\) together with the ordered partitioning
implied by the node sets of the process DAGs gives rise to the layered
DAG \(L^* = (G_{1}^{*} \oplus \dots \oplus G_{K}^*, \mathcal{L})\) with \(\mathcal{L} = (V_1,
\dots V_{10})\). Note that \(L^*\) and \(L^0\) are both layered DAGs
on the same vertex set \(V\) with the same layering \(\mathcal{L}\).
More importantly, the layer-induced subgraphs coincide, i.e., \(L_s^* =
L_s^0\) for all \(s \in [10]\).

Unfortunately, due to their complex physical nature, available process knowledge does not extend to how the $K$ processes  influence each other. In other words, the edge sets of \(L^*\) and \(L^0\) likely differ. However, Proposition \ref{prop:same_causal_orders} suggests that the process knowledge we gathered puts us in a favorable situation: Irrespective of whether there exist edges between layers, the set of all possible causal orderings is determined solely by the layers themselves, i.e. the process graphs. The complete layered DAG \(L^{\pi}\) for any \(\pi \in  \Pi_{L^{*}}\) is a super-DAG of the true layered DAG \(L^0\). Consequently, we may use common variable selection techniques from regression to prune \(L^{\pi}\) and infer the layered ground truth DAG \(L^0\) \citep[Section 2.5]{Buhlmann2014}.

\subsection{Sparse additive models (SpAM) for cross-process pruning}


We perform the pruning step by regressing each variable \(X_{v}\) onto its parent variables $X_k$, \(k\in pa_{L^{\pi}}(v)\).
The resulting active sets of predictors determine the final edge set, forming the estimate \(\hat{L}\).
Unless stated otherwise, the parent sets considered below are obtained from \(L^{\pi}\). When constructing \(L^\pi\) we add edges between the nodes from distinct processes $V_s$ and $V_t$, and leave the layer-induced subgraphs as they are. Similarly, we only consider edges between processes for pruning.  To prune,
we use the flexible nonparametric  framework of SpAMs \citep{Ravikumar2009}.
In a SpAM, the regression function for a variable $X_v$ with $v$ in layer $V_t$, is modeled as
\begin{equation*}
    f_v(x_{pa(v)}) = \mathbb{E}(X_v \mid X_{pa(v)}=x_{pa(v)})
    \approx \sum_{k\in pa(v)}f_{v,k}(x_k).
\end{equation*}
Sparsity is induced through a functional group-lasso penalty 
based on \(\lVert{f_{v,k}}\rVert_2 := \sqrt{\mathbb{E}(f_{v,k}(X_k)^2)}\), the \(L^2(P_{X_k})\) norm of component \(f_{v,k}(X_k)\).
For some \(\lambda \geq 0\), the resulting population-level minimization problem can be stated as:
\begin{equation*}
    \min_{f_{v,k} \in \mathcal{F}_{v,k}, \ k \in pa(v)} \left\{\frac{1}{2} \mathbb{E}\big(X_v - \sum_{k\in pa(v)}f_{v,k}(X_k)\big) + \lambda \sum_{k\in V_{1:t}} \lVert{f_{v,k}}\rVert_2\right\}.
\end{equation*}
Depending on the function classes $\mathcal{F}_{v,k}$, the minimization in SpAM becomes a convex program. Indeed, in practice, we express each function as \(f_{v,k}(x_k) = \sum_{l=1}^{p_{vk}} \psi_{vkl}(x_k) \beta_{vkl}\), where \(\{\psi_{vkl}\}_{l=1}^{p_{vk}}\) is a set of cubic splines.
Let \(\Psi_{vk}(x_k)\) be the \(1\times p_{vk} \) row vector of evaluations of the \(\psi_{vkl}\), and gather the \(\beta_{vkl}\) in the coefficient vector \(\theta_{vk} = (\beta_{vk1}, \ldots, \beta_{vk p_{vk}})^T\). Then \(f_{v,k}(x_k) = \Psi_{vk}\theta_{vk}\).
For estimation from data, we solve the empirical version of the SpAM problem, with cross-validation for tuning parameter selection. The set \(\{(k,v): k\in V_{1:t}, \hat{\theta}_{vk} \neq 0\} \cupdot pa_{L_s^*}(k)\) then marks the indices of the parent variables of $X_v$ remaining after pruning. Algorithm~\ref{alg:spam_search} outlines our procedure.

\begin{algorithm}
    \caption{SpAM cross-process learning}\label{alg:spam_search}
    \KwIn{Data \(\mathcal{D} \subseteq \mathbb{R}^p\), \  completed layered DAG \(L^\pi\),
        \ regularization parameter \(\lambda \geq 0\)}
    \KwOut{Edge set \(E_{\text{across}}\subset V\times V\).}
    \(E_{\text{across}} \gets \{\}\);\\
    \For{\(t = 2, \dots, K\)}{
        Form a causal ordering $\pi_t$ of the layer-induced subgraph \(L_t^\pi\);\\
        \For{\(v\) \textrm{along} $\pi_t$}{
            Obtain $(\hat \theta_{vk})_{V_{1:t}}$ through the following minimization problem:
            \begin{equation*}
                \min_{\theta_{vk}, \ k \in pa(v)} \left\{ \frac{1}{2|\mathcal{D}|}\sum_{x\in\mathcal{D}}
                \big(x_v - \sum_{k\in pa(v)} \Psi_{vk}(x_k)\theta_{vk} \big)^2+ \lambda \sum_{k\in V_{1:t}} \sqrt{\frac{1}{|\mathcal{D}|}\sum_{x\in\mathcal{D}} 
                    (\Psi_{vk}(x_k) \theta_{vk})^2} \right\}
            \end{equation*}
            \(E_{\text{across}} \gets E_{\text{across}}\cup\{(k,v): k\in V_{1:t}, \hat{\theta}_{vk} \neq 0\}\)
        }
    }
\end{algorithm}

For some variables \(X_v\) with $v$ in a layer \(V_t\), \(t \in \{2,\ldots,K\}\), the process experts are able to state functional relations, up to noise. Denote \(P_v^* := pa_{L^*}(v)\cap V_t\). Functional process knowledge takes the form $X_v\approx f^*_v(X_{P_v^*})$. To incorporate such knowledge, we alter Algorithm \ref{alg:spam_search} by inserting the known function.  The optimization problem is then
\begin{equation*}
    \min_{\theta_{vk}, \ k \in pa(v)} \left\{ \frac{1}{2|\mathcal{D}|}\sum_{x\in\mathcal{D}}
    \big(x_v - f^*_v(x_{P^*_v}) - \sum_{k\in V_{1:t}} \Psi_{vk}(x_k)\theta_{vk} \big)^2+ \lambda \sum_{k\in V_{1:t}} \sqrt{\frac{1}{|\mathcal{D}|}\sum_{x\in\mathcal{D}} 
        (\Psi_{vk}(x_k) \theta_{vk})^2} \right\}
\end{equation*}
Note  that Algorithm~\ref{alg:spam_search} admits a natural way to also
incorporate cross-process domain knowledge simply by adding the corresponding edge to \(E\) or dropping it from the corresponding parent set.

\begin{figure}[!ht]
    \includegraphics[clip, trim=0cm 2.8cm 0cm 3cm,width = \linewidth]{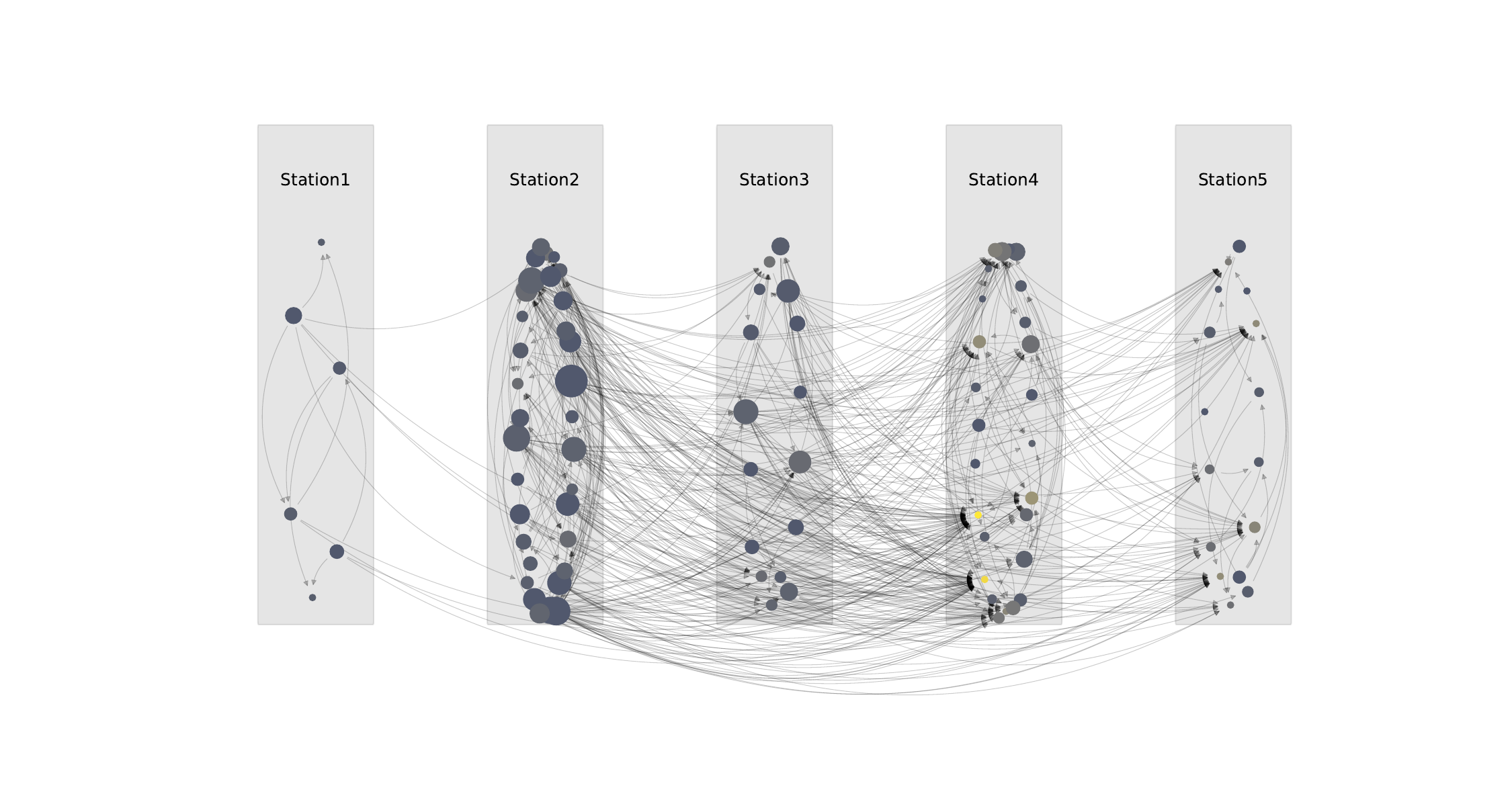}
    \centering
    \caption{Assembly line ground truth after edges have been learned
        between processes using SpAM. We depict production stations, not the processes they are decomposed in. Node size increases with the number of out-edges. Node color gets brighter with the number of in-edges. In terms of sparsity, the assembly line ground truth graph accounts for around \(10.2 \%\) of all possible connections.}
    \label{fig:assemblyline_overview}
\end{figure}
Figure \ref{fig:assemblyline_overview} illustrates the process graph \(\hat{L}\) after the pruning step using Algorithm \ref{alg:spam_search}. In the absence of latent confounding the sparsistency property of the SpAM procedure suggests that asymptotically we recover the true layered DAG \(L^0\). Even in the presence of confounding the synthetization procedure illustrated in the next section assures that data will be ``Markovian''. From here on we consider the layered DAG \(\hat{L}\) as ground truth in \texttt{causalAssembly}.

\subsection{Synthetization}\label{sec:drf}

To ensure that the semisynthetic data come from a distribution that factorizes according to the ground truth DAG \(\hat{L}\), we explicitly estimate the conditional distributions under the factorization implied by the independence constraints in \(\hat{L}\). To that end, we
employ distributional random forests (DRFs) \citep{Cevid2022} to obtain
estimates \(\hat{P}^{\text{DRF}}_X(X_v\mid X_{pa(v)})\), \(v \in V\),
along a causal ordering of \(\hat{L}\).

DRFs build on random forests \citep{Breiman2001} where the original splitting criterion is replaced with a distributional metric (in our case the maximal mean discrepancy (MMD) statistic \citep{Gretton2006}) in order to ensure homogeneity in each of the tree's leaf nodes. These give rise to neighborhoods of relevant training data points described by corresponding weighting functions. The average of the tree-specific weighting functions can then be used to estimate the conditional distributions. The following paragraphs give a brief overview of the DRF procedure.

\paragraph{DRF growing} Let \(L\) be a layered DAG. %
Assume we have grown \(N\) trees \(\mathcal{T}_1, \dots, \mathcal{T}_N\) and denote \(\mathcal{L}_j(x)\) the set of training data points that are assigned to the same leaf node as the test data point \(x\) in tree \(\mathcal{T}_j\). \cite{Cevid2022} define the weighting function by the average of the tree-specific weighting functions i.e. \(w_x(x_i) = 1/N \sum_{j=1}^N \mathbbm{1}(x_i \in \mathcal{L}_j(x))/\lvert \mathcal{L}_j(x) \rvert\). Let \(n\) be the sample size of the training data. Estimating the conditional distribution \(P(X_v \mid X_{pa_L(v)} = x)\) boils down to assigning weights to the point mass \(\delta_{x_{iv}}\) at \(x_{iv}\) in the empirical distribution of \(X_v\), i.e.
\begin{equation*}
    \hat{P}(X_v \mid X_{pa_L(v)} = x) = \sum_{i=1}^n w_x(x_i) \cdot \delta_{x_{iv}}.
\end{equation*}


Given \(\hat{P}^{\text{DRF}}_X(X_v\mid X_{pa(v)})\)
we sample a new dataset $\mathcal{D}_{\text{synth}}\subset\mathbb{R}^p$ from the production line as outlined in Algorithm \ref{alg:drf_sampling}.
By construction, \(\mathcal{D}_{\text{synth}}\) is drawn from the estimated joint distribution \(\hat{P}^{\text{DRF}}_X\) and is guaranteed to be Markov with respect to the ground truth layered DAG \(\hat{L}\).

\begin{algorithm}\label{alg:sampling}
    \caption{Sampling from the causal DRFs}\label{alg:drf_sampling}
    \KwIn{Finite dataset \(\mathcal{D}_{\text{original}}\subset
    \mathbb{R}^{p}\), (layered) DAG \(L\), number of samples \(n\) to be generated}
    \KwOut{A dataset \(\mathcal{D}_{\text{synth}} \subset
    \mathbb{R}^{p}\) of size $n$}
    $\mathcal{D}_{synth}\gets\emptyset$;\\
    Form a causal ordering $\pi$ of $L$;\\
    \While{$|\mathcal{D}_{\text{synth}}|<n$}{
    \For{\(v\) \textrm{along} $\pi$}{
    \eIf{\(pa_{L}(v) = \emptyset\)}{
    sample $\hat x_v\in\mathbb{R}$ via smooth bootstrapping from
    \(\{x_v\in\mathbb{R}:
    x\in \mathcal{D}_{\text{original}}\}\);}{
    sample $\hat x_v\in\mathbb{R}$ from $
        \hat{P}^{\text{DRF}}_X(X_v\mid
        X_{pa_{L}(v)} = \hat{x}_{pa_{L}(v)})$;
    }
    $\mathcal{D}_{synth}\gets\mathcal{D}_{synth}\cup\{\hat x_v\}$;
    }
    }
\end{algorithm}

\citet{Gamella2022} first proposed this approach using \citet{Sachs2005} data. By construction, it resolves confounding issues in the real data at the expense of \(\hat{L}\) potentially containing more cross-process edges than \( L^{0} \). 

\begin{remark}
    We emphasize the crucial role of the acquired process knowledge. Although the synthetization procedure is technically capable of fitting DRFs to any configuration of conditional distributions given a sufficient sample size, domain knowledge plays a pivotal role in ensuring the generated data closely resembles the real data. To illustrate this point, we conduct experiments in Section \ref{sec:initial_influence} where we assume no prior domain knowledge and instead rely solely on structure learning. Upon applying the semisynthetic procedure to a resulting process graph, we observe that in a majority of experiments, the causal discovery algorithm that initially generated the process graph structure tends to exhibit superior performance. As we demonstrate in Section \ref{sec:benchmark}, this is not the case for our semisynthetic data generator. We investigate, whether learning the cross-process edges via SpAM makes learning these edges easier. However, Figure \ref{fig:within_between} in Section \ref{sec:initial_influence} suggests that this is not the case. The combination of extensive process knowledge and highly flexible nonparametric techniques employed in \texttt{causalAssembly} renders it virtually impossible for any causal discovery algorithm to unfairly "game" graph recovery results.
\end{remark}

To accommodate the analysis of data with varying dimensionality, we provide the option to isolate and generate data for any of the \(5\) production stations that form the natural subunits of the assembly line. To avoid creating unintended confounding we refit the DRFs and apply Algorithm \ref{alg:drf_sampling} to each production station independently. Lastly, we want to stress that SpAMs are merely a pruning method. Their application in \texttt{causalAssembly} is not related to ANMs that have been discussed in Section \ref{sec:anms}.

\subsection{Semisynthetic data description}
\begin{figure}
    \centering
    \includegraphics[width=\textwidth]{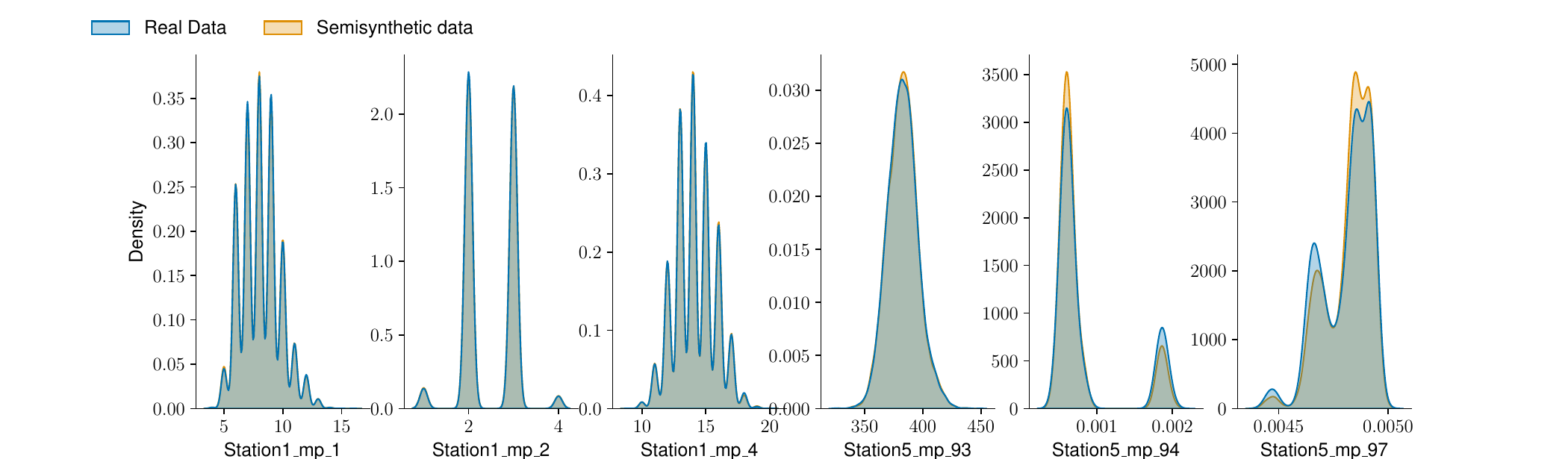} \\
    \caption{Kernel density plots (upper panel) of the same variables in the real (blue) and semisynthetic (yellow) data. Selection of those variables is based on comparing the highest (first three) and lowest (last three) agreement in terms of Kolmogorov-Smirnov statistic. 
    }
    \label{fig:marginal}
\end{figure}
To demonstrate that the semisynthetic ``Markovian'' data obtained from \texttt{causalAssembly} preserves key characteristics of
the real data, we calculate the two-sample
Kolmogorov-Smirnov (KS) statistic \citep{Kolmogorov1933} between all
variables in the real and a semisynthetic dataset. Figure~\ref{fig:marginal} shows kernel density estimates of variables with largest and smallest KS statistics among all non-source nodes. Even in unfavorable cases where the absolute difference between the marginal distributions is largest, the semisynthetic data only shows moderate deviations from the real data. To further illustrate the similarity of the semisynthetic data to the real data, we depict pairwise correlation agreement in Figure~\ref{fig:correlation_comparison} in the Appendix.

In Figure \ref{fig:scatter}, we depict bivariate
relationships between selected direct causes on the x-axis and their
effects on the y-axis according to the ground truth DAG \(\hat L\).
Variables were selected within stations in (a) and (b) and between stations in (c). The size of the parent sets for the
effect variables are $1,3,6$ for (a), (b), and (c), respectively. Naturally, with increasing parent set size bivariate
relationships become harder to discern. While (a) depicts a clear linear relationship between the selected measurements, the relationships in (b) and (c) are masked by the remaining parent variables.

\begin{figure}
    \centering
    \includegraphics[width=\textwidth]{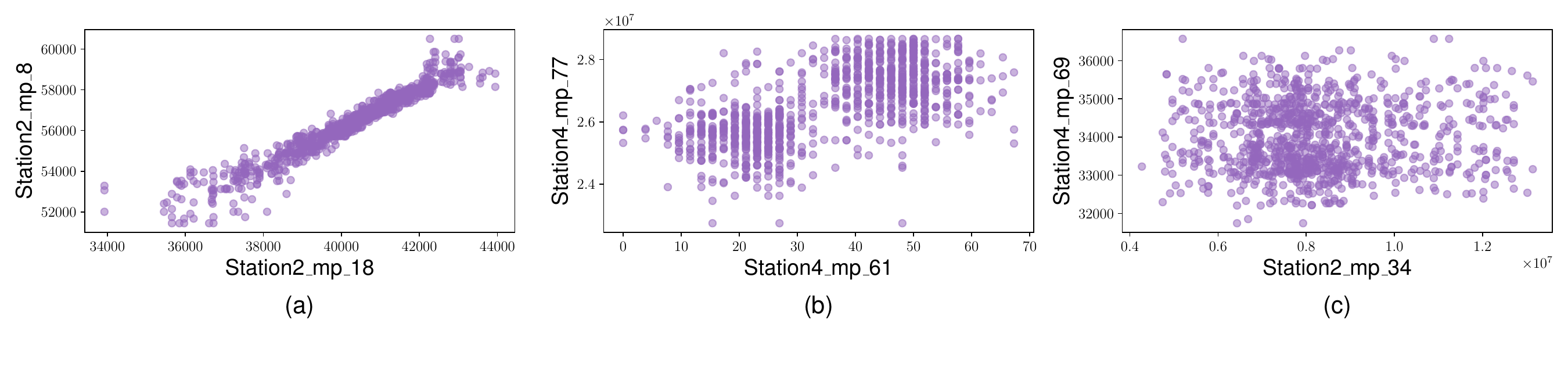}
    \caption{Bivariate scatter plots of selected nodes to showcase the different types of bivariate patterns to expect in data generated with \texttt{causalAssembly}. All node pairs are causally linked with x-axis variables being parents of y-axis variables. The number of parents of each y-axis variable varies from one parent in (a), three parents in (b) and six parents in (c).}
    \label{fig:scatter}
\end{figure}


\section{Benchmark results}\label{sec:benchmark}

As a proof of concept, we present first causal discovery benchmark results.
In order to make sure that privacy concerns are respected, we first apply the semisynthetization procedure to the real data from the manufacturing plant. We then sample once from the joint distribution in \texttt{causalAssembly}. The resulting data is used for benchmarking as shown below. We present further benchmark results based on the individual production stations in Section \ref{sec:addition_benchmarks}.

The purpose of the study we report on here is to demonstrate the versatility of the data generation scheme. A more elaborate set of experiments comprising varying sample size and adjusting hyperparameters will be taken up in future work. The PC algorithm does not output a DAG. Thus, to compare results additional harmonization steps are necessary. We implement two strategies \textit{CPDAG-transform} and \textit{average-dag} outlined in Section \ref{sec:alg_metrics}.
\begin{figure} 
    \includegraphics[width = \linewidth]{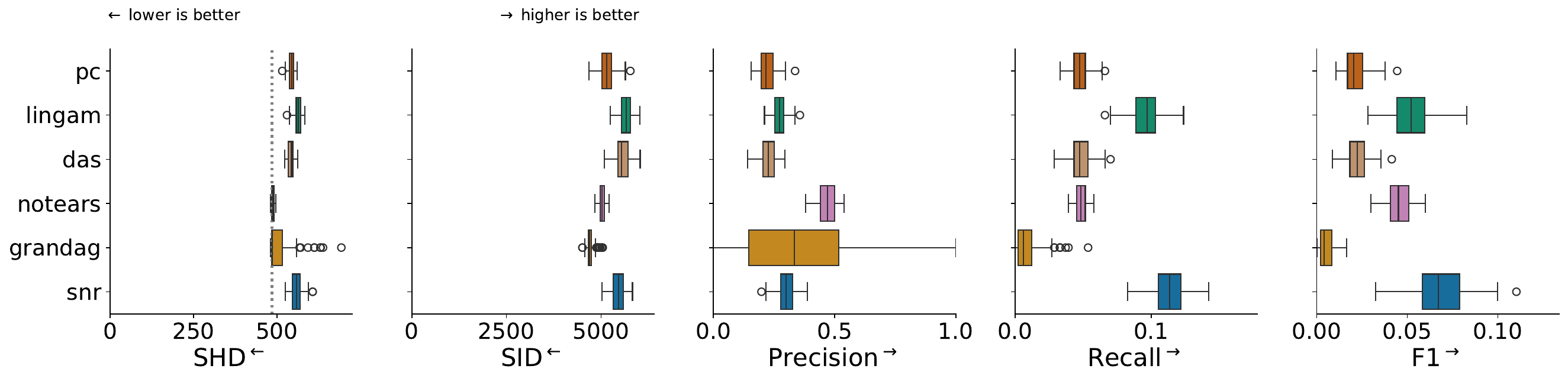}
    \centering
    \caption{Chosen metrics for assessing causal discovery of the full assembly line (\(\lvert V \rvert = 98\)) using \(100\) simulation runs on a sample size of \(n = 5000\). The dashed line corresponds to the SHD between the empty graph and ground truth indicating the number of edges in the ground truth graph. Note that we use DAS rather than SCORE on the full line due to its improved scaling to higher dimensions.}
    \label{fig:full_benchmark}
\end{figure}
As the results following these strategies do not seem to differ substantially, we report the \textit{average-dag} results and refer to Appendix~\ref{sec:app_c} for the remaining experiments. A brief overview of the algorithms and metrics is provided in Section~\ref{sec:alg_metrics}. We standardize the data sampled in each of the \(100\) simulation runs before structure learning and report the average varsortability in Table~\ref{table:varsort} in Section~\ref{sec:app_c}. The sortnregress (snr) routine serves as a reference point, as on standardized data it corresponds to choosing some ordering at random and regressing each node on its ancestors (\textit{random regress}) \citep[see also][]{Reisach2021}.

\begin{figure}
    \includegraphics[width = \linewidth]{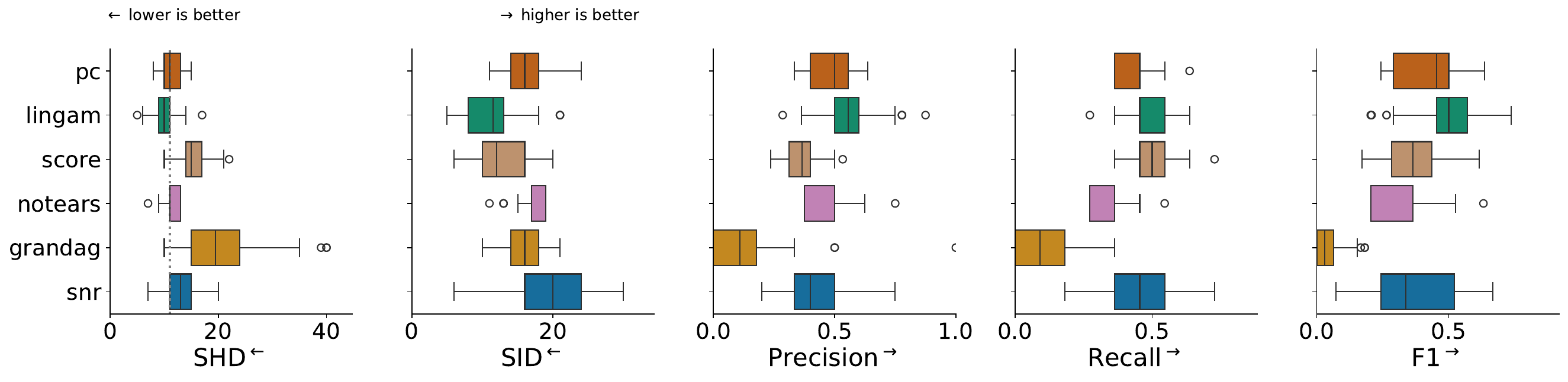}
    \centering
    \caption{Chosen metrics for assessing causal discovery outcomes on Station \(3\) (\(\lvert V_5 \cup V_6 \rvert = 16\)) using \(100\) simulation runs on a sample size of \(n = 500\). The dashed line corresponds to the SHD between the empty graph and ground truth indicating the number of edges in the ground truth graph.}
    \label{fig:station3_benchmark}
\end{figure}

Figures~\ref{fig:full_benchmark} and~\ref{fig:station3_benchmark} illustrate the benchmark results for the full line and Station \(3\), respectively. We focus on Station \(3\) because its results differ significantly from those of the full line. The benchmark results for the remaining four process stations are displayed in Section~\ref{sec:app_c}.

Regarding the full line benchmark results in Figure \ref{fig:full_benchmark}, the default methods fail to beat an empty graph in terms of structural Hamming distance (SHD). Overall, NOTEARS performs best in terms of SHD and SID, but this is partly due to its tendency to output very sparse graphs, as evidenced by low Recall scores. Linear methods such as DirectLiNGAM and the PC algorithm with standard partial correlation tests perform similarly to \textit{random regress}. GraN-DAG exhibits higher variability in results, with highly variable precision and very low recall and F1 scores.
Inspecting the results from Station \(3\) depicted in Figure \ref{fig:station3_benchmark}, we observe that most selected algorithms outperform random regress, especially in terms of structural intervention distance (SID) and F1 score. DirectLiNGAM exhibits the best performance across most metrics, while GraN-DAG again struggles with precision, recall, and F1. The PC algorithm and SCORE perform similarly, and NOTEARS fails to beat \textit{random regress} in this case.

We emphasize that these results are based on out-of-the-box implementations of the algorithms, with no attempt to tune them to the data. Additionally, all the considered methods ignore the prior knowledge implied by the layered structure of the processes. We expect that causal discovery tools that can efficiently incorporate the layered structure will achieve better performance. Furthermore, tuning the hyperparameters of the algorithms and using statistical methods that are better equipped to handle complex nonlinear relationships is likely to yield better results.

While limited, this first benchmark study reveals that standard off-the-shelf methods struggle with data generated by \texttt{causalAssembly}. This largely corroborates findings from empirical studies, which show that the learning accuracy reported in the literature on synthetic data overestimates performance on more complex datasets \citep{Constantinou2021,Rios2022}.


\section{Conclusion and Limitations}\label{sec:conclusion} 



Empirical validation is key in understanding how causal discovery algorithms perform outside their laboratory settings. In this work, we leverage a complex dataset comprising measurements from a highly automated assembly line in a manufacturing plant, for which we are able to put together partial ground truth causal relationships on the basis of a detailed study of the underlying physics. Based on this dataset, we use the associated ground truth and develop a versatile synthetization pipeline based on DRFs that helps overcome common real world data problems such as lack of ground truth, confounding, and privacy concerns. At the same time, by preserving the real data characteristics as much as possible while making sure that the observational distribution is Markov with respect to the obtained ground truth, we overcome the issue of purely synthetic data often being too simplistic and not generalizable to real world settings. The resulting data generator can be used to benchmark causal discovery algorithms based on complex production data. We demonstrate this by providing initial benchmarking results for some well-known causal discovery algorithms.

An important limitation is that currently \texttt{causalAssembly} supports the generation of continuous data only. However, the methodology presented is more general and supports mixed data. We plan to include discrete features by selecting a wider range of processes in future releases of \texttt{causalAssembly}. Another limitation comes from the fact that ground truth relations are only available within and not across processes. In future releases we plan to address this by continuously validating cross-process edges together with process experts. Lastly, it would be interesting to take advantages of opportunities to involve interventional samples.





\acks{We thank the anonymous reviewers and Jens Ackermann for helpful comments and insights. The research for this project was partly conducted by authors employed at Robert Bosch GmbH.
}

\clearpage
\newpage
\bibliography{bib}

\clearpage
\newpage
\appendix

\section{Assembly line dataset}\label{sec:app_a}

We leverage a complex production dataset comprising real measurements from a highly automated assembly line in a manufacturing plant. The assembly line consists of multiple production stations along which individual components are joined together. Parts are processed in these production stations and are sequentially passed to the next station according to the line structure.

Often, stations perform the same process several times either in succession or in parallel to reduce overall cycle time. Raw components are added to the assembly at the respective stations. The quality of the assembly process depends on the respective machine state and settings (e.g. tool state, control parameters), the environmental conditions, and on the raw state of the components or intermediates. We carefully study the physical processes that take place in each of the stations.

\subsection{Mapping process knowledge}\label{sec:process_knowledge}

\textbf{Press-in \quad} In order for the inner cylindrical component to be pressed into the bore, mechanical force has to exceed friction force. The friction force depends on component properties, such as the geometrical dimensions (especially diameters and axial lengths), structural stiffness as well as the surface conditions (e.g. surface roughness). Also, machine properties such as misalignment of the pressing machine tool with respect to the work piece holder and cylindrical bore axis can affect the process. None of these properties is measured directly for each component. Depending on the product type, the nominal values of the component properties may differ.

\begin{table}
    \centering
    \caption{Description of relevant force and position features extracted from the measurements during the press-in process.}
    \tymin=50pt
    \begin{tabulary}{1.0\textwidth}{L J}
        \toprule
        Parameter           & Description                                                                                                                                                                                                                                                                                                                                                                                                                                                                                  \\ \midrule
        \(F_{\text{offs}}\) & Average force measured during closing of the gap between tool and work piece. This depends on the acceleration and mass of the tool as well as the force sensor measurement error.                                                                                                                                                                                                                                                                                                           \\ \addlinespace
        \(F_1\)             & Press-in force (friction force) required to reach the block position (\(S_1\)) as extracted from gradient detection algorithm.                                                                                                                                                                                                                                                                                                                                                               \\ \addlinespace
        \(F_2\)             & Maximum force at the end of the process. This force depends on the structural stiffness and the difference between the positions \(S_2\) and \(S_1\).                                                                                                                                                                                                                                                                                                                                        \\ \addlinespace
        \(S_0\)             & Axial position of tool at first force increase as extracted from gradient detection algorithm. Position depends on geometrical tolerances of the machine and components as well as on the calibration of the position sensor.                                                                                                                                                                                                                                                                \\ \addlinespace
        \(S_1\)             & Axial position of tool at reaching the block position. Position depends on the relative displacement \(S_0\). \(S_1\) is affected by the length of the bore as well as by the elastic deformation of the machine. Deformation is influenced by the required press-in force \(F_1\) and by the structural machine and component stiffness. Once this gradient position is detected by the process control unit during process execution the command for reversing the tool movement is given. \\ \addlinespace
        \(S_2\)             & Maximum tool position where the tool direction of movement is effectively reversed. This is affected by the position \(S_1\), by the time required for computation of the gradient point, the tool speed, by the inertial mass of the tool and by the effective structural stiffness.                                                                                                                                                                                                        \\
        \bottomrule
    \end{tabulary}%
    \label{table:vardesc_pressin}\\
\end{table}%

\begin{figure}
    \includegraphics[width=.5\textwidth]{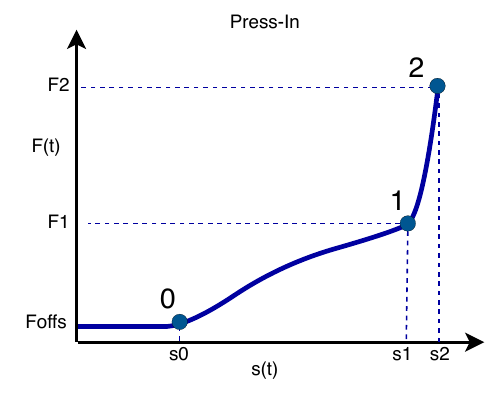}
    \centering
    \caption{A typical Force (y-axis) and Position (x-axis) curve during press-in illustrating the characteristics of the process.}
    \label{fig:force_displacement}
\end{figure}

During the fitting process the force acting on the pressing machine tool and its current vertical position are measured jointly using special sensors. In order to control and monitor the process both signals are analyzed during and after the process by the process control unit, for example using gradient detection algorithms. The change in measured position is the sum of both the elastic deformation of the pressing machine (driven by the force and structural machine stiffness) and the actual movement of the inner component relative to the bore. Figure \ref{fig:force_displacement} demonstrates how the machine tool force behaves along the vertical position. The most important features extracted from this procedure are stored in the manufacturing execution system. Table \ref{table:vardesc_pressin} depicts a selection of these and lays out dependencies.

\begin{figure}
    \includegraphics[width=.5\textwidth]{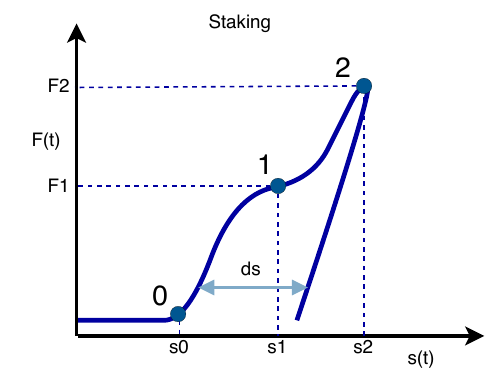}
    \centering
    \caption{A typical Force (y-axis) and Position (x-axis) curve during staking illustrating the characteristics of the process.}
    \label{fig:force_displacement_staking}
\end{figure}

\textbf{Staking \quad} In staking the deformation process of the material is mainly driven by two factors. One factor is the material yield strength and hardening behavior. The other factor is the compression stress. The latter is imposed by the force acting on the tool and by the contact area between tool and component. If several staking processes are happening on the same component with several bore positions, the unknown material state is likely to be similar, confounding the observed force values. Again a typical force and position curve of a staking process as well as a description of extracted features including their interrelations are demonstrated in Figure \ref{fig:force_displacement_staking} and Table \ref{table:vardesc_staking}, respectively.


\begin{table}
    \centering
    \caption{Description of relevant force and position features extracted from the measurements during the staking process.}
    \tymin=50pt
    \begin{tabulary}{1.0\textwidth}{L J}
        \toprule
        Parameter      & Description                                                                                                                                                                                                                                                                                                                                            \\ \midrule

        \(F^*_1\)      & Staking force required to reach the turning point of increasing structural stiffness  as extracted from the detection algorithm. Once this turning point is detected by the process control unit during process execution the process command for reversing the tool movement is given after an additional, pre-defined force delta has been exceeded. \\ \addlinespace
        \(F^*_2\)      & Maximum force at the end of the process. This force depends on the structural stiffness and the difference between the positions \(S^*_2\) and \(S^*_1\).                                                                                                                                                                                              \\ \addlinespace
        \(S^*_0\)      & Axial position of tool at first force increase as extracted from gradient detection algorithm. Position depends on geometrical tolerances of the machine and components as well as on the calibration of the position sensor.                                                                                                                          \\ \addlinespace
        \(S^*_1\)      & Axial position of tool at reaching the turning point. Position depends on the relative displacement \(S_0\). \(S_0\) is affected by the length of the bore as well as by the elastic deformation of the machine. Deformation is influenced by the required press-in force \(F^*_1\) and by the structural machine and component stiffness.             \\ \addlinespace
        \(S^*_2\)      & Maximum tool position where the tool direction of movement is effectively reversed. This is affected by the position \(S^*_1\), by the time required for computation of the turning point, the tool speed, by the inertial mass of the tool and by the effective structural stiffness.                                                                 \\ \addlinespace
        \(\Delta S^*\) & Difference in axial position of the tool after removing the pressing force measured at a pre-defined remaining force value. This indicates the true deformation length in the bore without the elastic machine deformation.                                                                                                                            \\
        \bottomrule
    \end{tabulary}%
    \label{table:vardesc_staking}\\
\end{table}%

Often, press-in and staking processes are run on the same machine. Unobserved machine properties such as structural stiffness, tool misalignment and force sensor positions may impact both processes. Furthermore, confounding can be caused by similarities of the component properties (same component with several bores) or due to batch effects.

\begin{figure}
    \includegraphics[width = \linewidth]{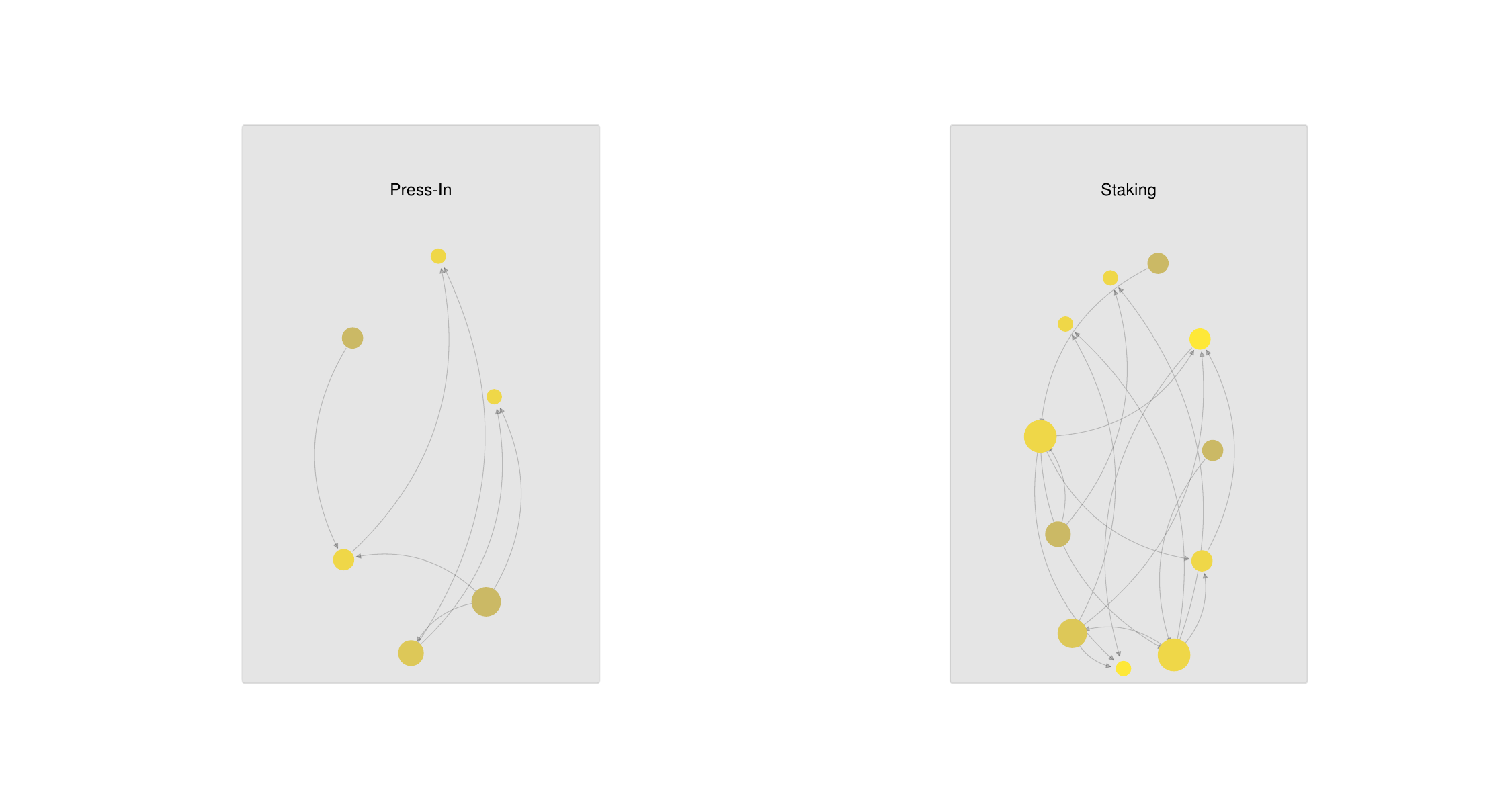}
    \centering
    \caption{Process DAGs of an exemplary press-in and staking process. Node size increases with the number of out-edges. Node color gets brighter with the number of in-edges.}
    \label{fig:pressing_staking_example}
\end{figure}

Figure \ref{fig:pressing_staking_example} depicts an exemplary press-in and staking process. Depending on the production cell and process step these might vary slightly in size and complexity along the production line. From Figure \ref{fig:pressing_staking_example} we can see that the staking process is larger and more complex than the press-in process. This reflects the fact that a staking process is an extension of a press-in process.

\section{Proof of Proposition \ref{prop:same_causal_orders}}\label{sec:proofs}

\begin{proof}
    It suffices to show  \(\Pi_L \subseteq \Pi_{L^{\prime}}\).   So, let \(\pi \in \Pi_L\) and consider any two nodes $v,w\in V$.  If $v\in V_s$ and $w\in
        V_t$ with $s\neq t$, then we must have $s<t$ because $L$ is a layered DAG.  Since $L'$ is based on the same partition $\mathcal{L}$, we have $\pi\in\Pi_{L'}$.
    If instead $v,w\in V_s$ for some $s\in [K]$, then $L_s=L^\prime_s$.  We conclude that $v$ is a descendant of $w$ in $L$ if only if this is the case in $L'$.  Thus, again $\pi\in\Pi_{L'}$.
\end{proof}
\section{Semisynthetic data generator}\label{sec:app_b}

The semisynthetic data generation tool is available through the Python library \texttt{causalAssembly} (\url{https://github.com/boschresearch/causalAssembly}). In the library, we also provide the semisynthetic data itself (and for convenience one set of datasets with sample size $n=500$) as well as the partial ground truth. All benchmark results presented both in the main text and the supplementary sections will also be accessible through the library.

\subsection{Quality of the DRF procedure}

To avoid overfitting when learning DRFs, we grow \(N = 2000\) trees on a random subset of the training data for each conditional distribution. When forming each tree, the subsets are split again; one part is used to grow the tree, the other part is used to populate the leave nodes. We choose the Gaussian kernel for the MMD splitting criterion with bandwidth set according to the \textit{median heuristic} \citep{Gretton2012}. Table \ref{table:tuning} in Section \ref{sec:tuning} gives an overview over the all relevant hyperparameters.

\begin{figure}
    \includegraphics[width = .8\linewidth]{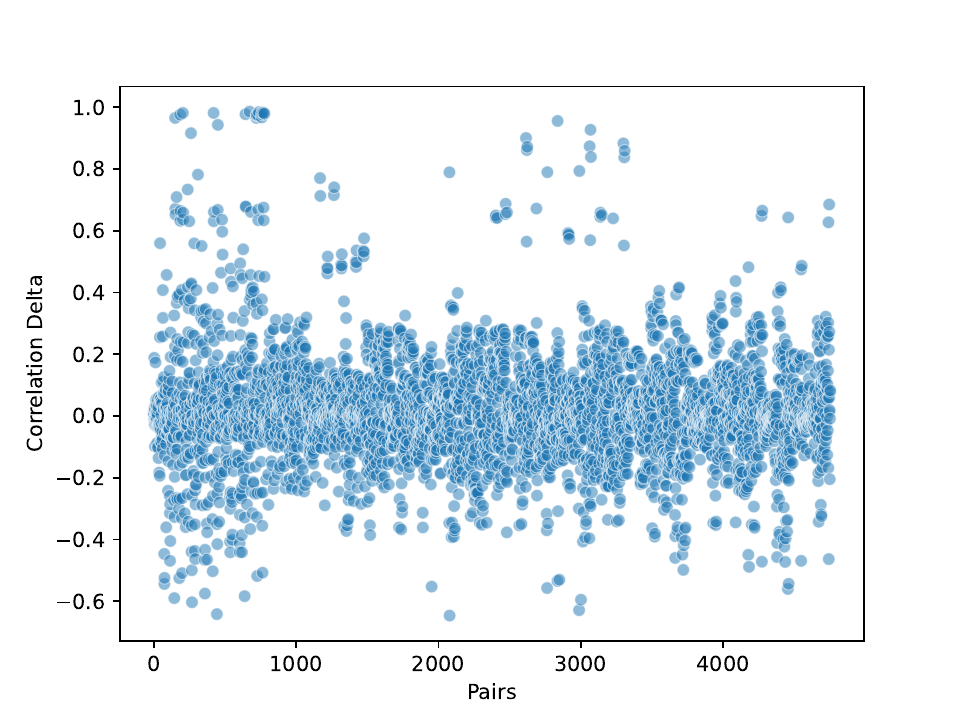}
    \centering
    \caption{Pairwise correlation agreement between semisynthetic and real data. We depict the difference between the pearson correlation of the same pair in one draw of the semisynthetic data and the real production line data. Sample size is \(n = 15.581\) for semisynthetic and real data.}
    \label{fig:correlation_comparison}
\end{figure}

In the interest of quality control, we compare correlation patterns in the synthetic data largely agree with those of the real data as shown in Figure \ref{fig:correlation_comparison}. Some divergence is to be expected as the real data comes with a wide range of confounding problems which our procedure deals with by employing the DRF procedure described in Section \ref{sec:drf}.
Figure \ref{fig:correlation_comparison} confirms that for most of the pairs among the \(p = 98\) variables, correlation patterns agree.

\subsection{Influence of initial learning}\label{sec:initial_influence}

\begin{figure}
    \centering
    \includegraphics[width = \linewidth]{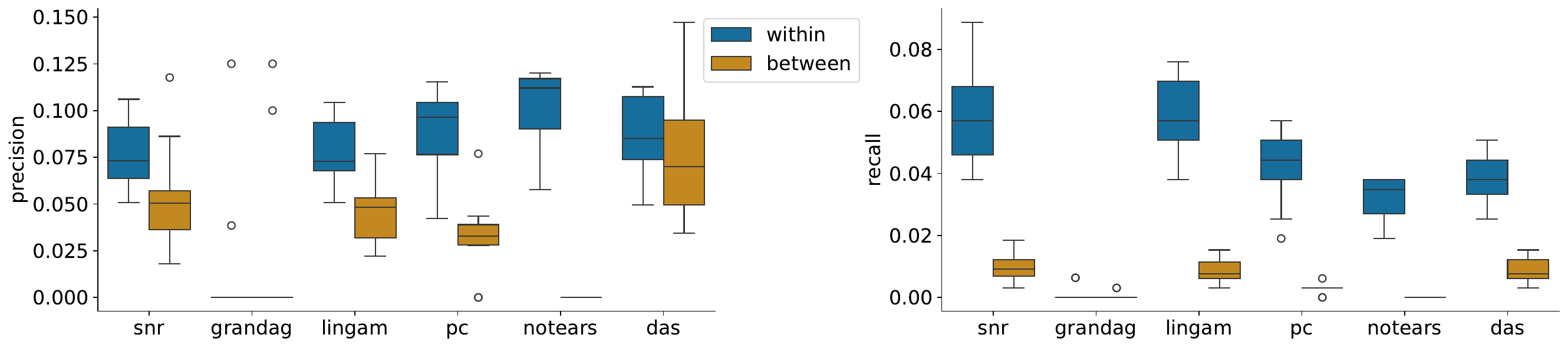}
    \caption{Comparison between within and cross process edges. Simulation run with \(50\) repetitions and sample size \(n = 5000\). Precision and recall scores are relative to the amount of within and cross-process edges, respectively.}
    \label{fig:within_between}
\end{figure}

Before investigating the influence of the absence of prior knowledge, we engage in a brief comparison of between and within process edge recovery. Figure \ref{fig:within_between} demonstrates that cross-process edges are indeed not recovered more easily as opposed to within-process edges. If any, the opposite is true and within-process edges achieve somewhat better precision and recall scores.

\begin{figure}
    \includegraphics[width = \linewidth]{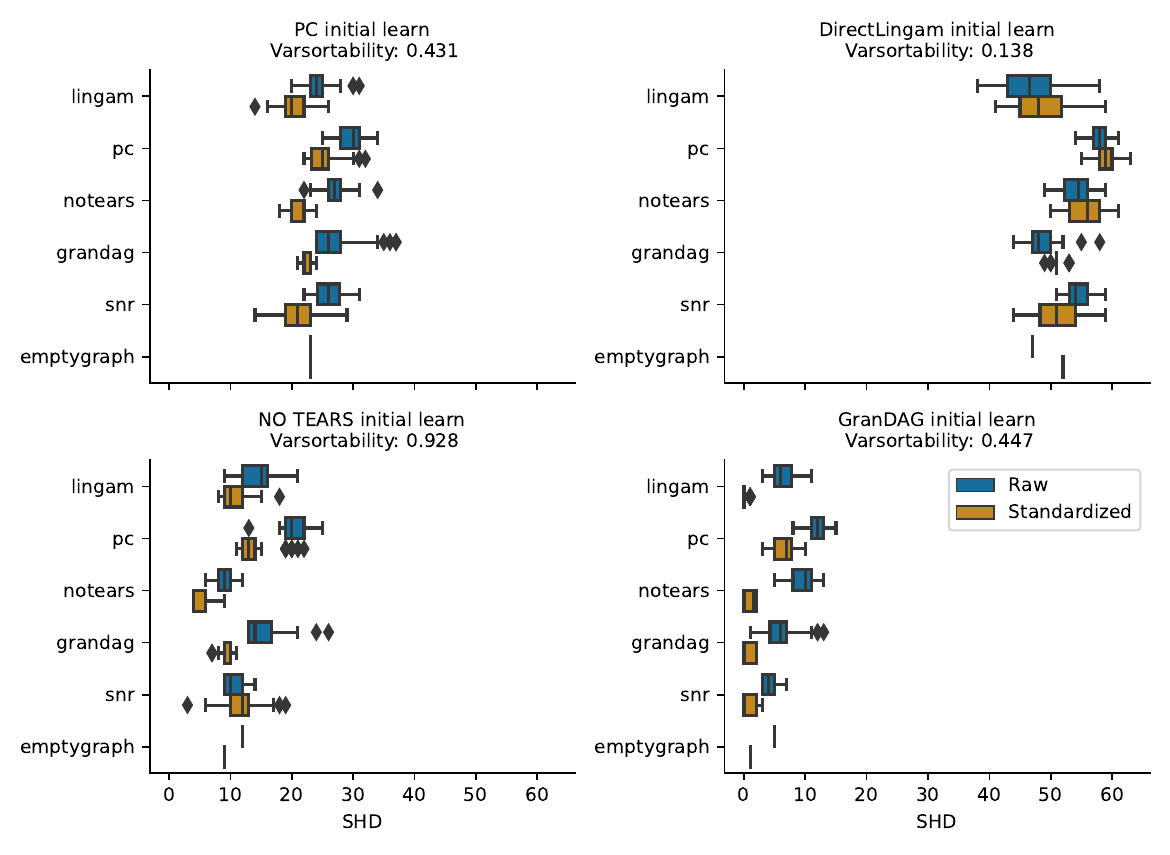}
    \centering
    \caption{SHD for initial learning based on chosen causal discovery algorithms. Simulation run with \(50\) repetitions.}
    \label{fig:initial_learning}
\end{figure}

\begin{figure}
    \includegraphics[width = \linewidth]{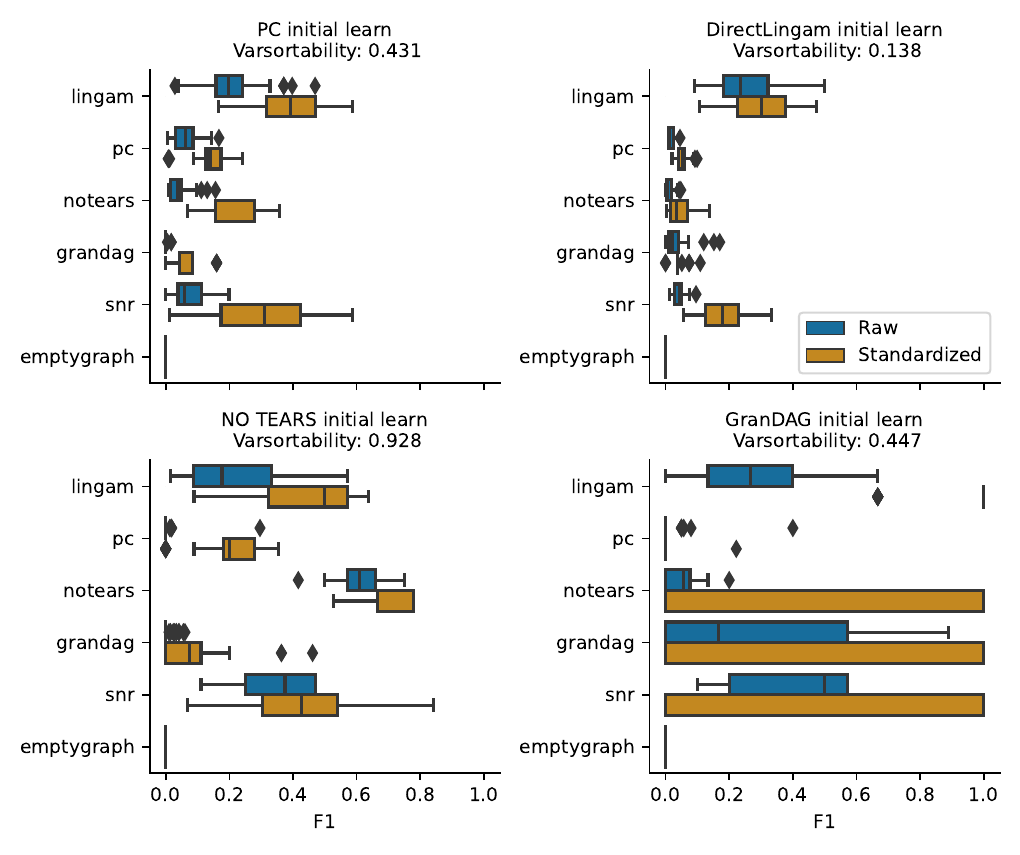}
    \centering
    \caption{\(F_1\) score for initial learning based on chosen causal discovery algorithms. Simulation run with \(50\) repetitions.}
    \label{fig:initial_learning_f1}
\end{figure}

Turning to the absence of prior knowledge, learning edges within and across processes using causal
discovery algorithms only without prior knowledge leaves distinct traces. Benchmarking outcomes are then
tilted in favor of the routine employed in this initial step. We
demonstrate this on processes that take place in Station \(3\) consisting of \(16\) nodes. We ignore existing process knowledge and learn from the data only. Figures \ref{fig:initial_learning} and \ref{fig:initial_learning_f1} report SHD and \(F_1\) score between the ground truth DAG learned by the initial causal discovery procedure and results based on corresponding semisynthetic data.

Next to overall patterns left by the initial learning step, we present sensitivity results with respect to measurement scale. The real data obtained from the process control unit carry a wide range of different measurement scales. These scales should ideally be irrelevant with respect to causal relationships. In a first step, we keep the raw data scale both for the initial learning and for the benchmark run based on the obtained ground truth. We report the average varsortability across simulation runs. In a second step, we standardize all data. We also report results for the
empty graph to keep track of the number of edges inferred during the initial graph learning step. If the SHD of the empty graph changes under standardization then data scale has an effect on the initial ground truth.

DirectLiNGAM, NOTEARS, and GraN-DAG show a clear performance edge when
the ground truth is originally obtained using the respective routine. This pattern is most clear in Figure \ref{fig:initial_learning_f1}. The PC algorithm does not seem to benefit too greatly which can to some degree be attributed to the random DAG completion. In contrast to the remaining methods, the PC algorithm is agnostic to measurement scale. DirectLiNGAM is in theory also scale agnostic, however scaling
may influence the adjacency matrix estimation via adaptive lasso when
predictors are not standardized. Regarding NOTEARS and GraN-DAG
measurement units play a substantial role. NOTEARS obtains a ground
truth with varsortability close to one on unstandardized data and
recovers a very sparse DAG on standardized data. This suggests that NOTEARS essentially orders variables according to their variances when
learned on raw data. GraN-DAG results are highly variable throughout
all initial learning methods. The instability becomes particularly
severe when applied to standardized data.

Overall, except for the PC algorithm there are clear traces left by initial learning algorithms applied to process cell structure learning discovery. In this context, the question arises whether to simply select a random DAG to represent process or station relationships. Following the proposed procedure and given sufficient sample size, the DRFs will be able to produce weights that allow drawing from such a ground truth graph. The choice of random DAG however influences strongly how close resulting synthetic data to the real world complexity will be. In the complete absence of ground truth knowledge, such a procedure might be an alternative to specifying a random SCM.

\section{Benchmarking}\label{sec:app_c}

\subsection{Algorithms and metrics}\label{sec:alg_metrics}

\paragraph{Algorithms} The following well-known structure learning algorithms are employed in our work.

\textit{PC algorithm} \citep{Spirtes1993} (\texttt{pc})\\
A constraint based algorithm that in a first step performs conditional
independence tests in a resource efficient way returning a skeleton
and in a second step orients as many edges as possible. We use the
stable version of the algorithm proposed by \cite{Colombo2014}.
Conditional independence tests are performed using Fisher
z-transformed (partial) Pearson correlation coefficients. Since the PC
algorithm outputs a completed partially directed acyclic graph
(CPDAG), we choose in each run a random DAG that is a member of the
MEC implied by the CPDAG.

\textit{DirectLiNGAM} \citep{Shimizu2011, Hyvarinen2013} (\texttt{lingam})\\
The direct method to learning the LiNGAM model \citep{Shimizu2006}
referred to in Section \ref{sec:identifiability}. First, the procedure
detects source variables by employing entropy-based measures to find
independence between the error variables. The \textit{most
    independent} variable is placed first in the causal ordering and its
effect removed from the other variables. As the ordering of the
residuals resembles that of the variables themselves the algorithm
continues to find the next variable in the ordering by finding the
next \textit{source}-residual.

\textit{NOTEARS} \citep{Zheng2018, Zheng2020} (\texttt{notears})\\
A score-based method that recasts the discrete DAG constraint as a
continuous equality constraint. This allows for the application of a
wide range of techniques for non-convex programs. In NOTEARS a linear
SCM is assumed and for each variable, a feed-forward neural network is
trained that gets all other variables as input. The graph structure is
then dynamically read-off from the networks during training by
aggregating the weights of the input layer of each network. Acyclicity
of the graph is enforced by training the networks simultaneously in an
augmented Lagrangian schema such that the matrix exponential of the
adjacency matrix has small trace.

\textit{GraN-DAG} \citep{Lachapelle2020} (\texttt{grandag})\\
A score-based method and an extension of NOTEARS. In contrast to
NOTEARS, the relation structure is read-off from the networks by
aggregating all the weights till the output layer of the neural
networks individually.

\textit{SCORE} and \textit{DAS} \citep{Rolland2022,Montagna2023} (\texttt{score, das}) \\
A order-based method for learning nonlinear additive Gaussian noise models which has been shown to be identifiable from observational data \citep{Peters2014a}. The method evolves around estimating the causal ordering based in approximating the score (the map \(\nabla \log p(x)\) for a differentiable density \(p(x)\)) of the data distribution. Under this model, leaf nodes indicate a constant entry in the associated diagonal element of the score's Jacobian. This allows for sequential leaf node identification implying a causal ordering. The method concludes with a pruning step to remove spurious edges. Its extension \textit{discovery at scale} (DAS) reduces the complexity of the pruning step.

\textit{sortnregress} \citep{Reisach2021} (\texttt{snr})\\
A diagnostic tool to expose data scenarios with high varsortability.
The procedure sorts variables by increasing marginal variance and then
regresses each variable on their ancestors adding an \(\ell\)-1 type
penalty to induce sparsity.

\paragraph{Metrics} The choice of evaluation metrics for assessing causal discovery outcomes matters \citep[see][]{Gentzel2019}. We employ a selection of common structural performance metrics and note that in a full scale benchmark these should be extended to include a greater variety of evaluation measures.

\textit{Precision, Recall and \(F_1\) score:} \\
Precision in the graphical context refers to the fraction of correctly
identified edges among all edges inferred. Recall, sets these
correctly identified edges in relation to all edges present in the
ground truth. The \(F_1\) score is the harmonic mean of Precision and
Recall, i.e. \(F_1 =
2\frac{\text{Precision}\cdot\text{Recall}}{\text{Precision}+\text{Recall}}\).

\textit{Structural hamming distance (SHD):} \\
The SHD counts how many edge insertions, deletions, and reversals are
necessary to turn the estimated DAG into the ground truth DAG.

\textit{Structural Interventional Distance (SID)} \citep{Peters2015}:\\
The SID counts the number of incorrectly inferred interventional distributions in the estimated DAG when compared to the ground truth DAG.

\paragraph{Harmonization} We employ two harmonization strategies \textit{CPDAG-transform} and \textit{average-dag}.

(i) We transform all graphs, including the ground truth DAG, to their completed partially directed acyclic graph (CPDAG) representing the associated MEC \citep[see][for details]{Andersson1997}. Metrics will be calculated on CPDAGs only. Note that the SID is defined for DAGs and will not be part of the evaluation metrics following \textit{CPDAG-transform}.

(ii) The second strategy, \textit{average-dag}, involves choosing one member DAG from the implied MEC of algorithms that do not output a DAG. We then enumerate all DAGs in the MEC, calculate their structural Hamming distance (SHD) with respect to the ground truth DAG, and take the one whose SHD is closest to the average SHD. When running experiments on the full line, enumerating DAGs in their MEC becomes computationally infeasible, so we choose one random DAG from the associated MEC.

\subsection{Additional benchmarking results}\label{sec:addition_benchmarks}

\begin{table}
    \caption{All benchmarks are run on a Linux machine with 32 GB Memory and 8 cores.}
    \label{table:compute}
    \centering
    \begin{tabular}{l c}
        \toprule
        Result                                                                                                        & Compute time \\ \midrule
        Full assembly line benchmark Fig \ref{fig:full_benchmark} and \ref{fig:full_cpdag_benchmark} in the main text & 16.90 hours  \\
        Station \(1\) benchmark Fig \ref{fig:station1_benchmark} and \ref{fig:station1_cpdag_bench}                   & 1.07 hours   \\
        Station \(2\) benchmark Fig \ref{fig:station2_benchmark} and \ref{fig:station2_cpdag_bench}                   & 4.39 hours   \\
        Station \(3\) benchmark Fig \ref{fig:station3_benchmark} and \ref{fig:station3_cpdag_bench}                   & 2.22 hours   \\
        Station \(4\) benchmark Fig \ref{fig:station4_benchmark} and \ref{fig:station4_cpdag_bench}                   & 3.54 hours   \\
        Station \(5\) benchmark Fig \ref{fig:station5_benchmark} and \ref{fig:station5_cpdag_bench}                   & 2.33 hours   \\
        \bottomrule
    \end{tabular}
\end{table}

\begin{table}
    \centering
    \caption{Average varsortability over $100$ draws using the semisynthetic data generation procedure for the full assembly line and for each production station.}
    \begin{tabular}{lr}
        \toprule
                  & Varsortability \\
        \midrule
        Full line & 0.486122       \\
        Station1  & 0.669167       \\
        Station2  & 0.348319       \\
        Station3  & 0.330588       \\
        Station4  & 0.247204       \\
        Station5  & 0.210417       \\
        \bottomrule
    \end{tabular}
    \label{table:varsort}
\end{table}

We present additional benchmarks for the CPDAG harmonization strategy and the remaining production stations from the MEC enumeration strategy. Average varsortability before standardization is reported in Table \ref{table:varsort} and Table \ref{table:compute} states compute time for each of the benchmark scenarios. As expected, varsortability ranges between $0.7$ in the first station and $0.22$ in station five. The full assembly line exhibits varsortabiliy close to $0.5$ indicating that ordering variables by their marginal variances does not give any insights regarding the causal ordering. In each simulation run, we sample from the DRF regrown on the corresponding layer-induced subgraphs.

\begin{figure}
    \includegraphics[width = \linewidth]{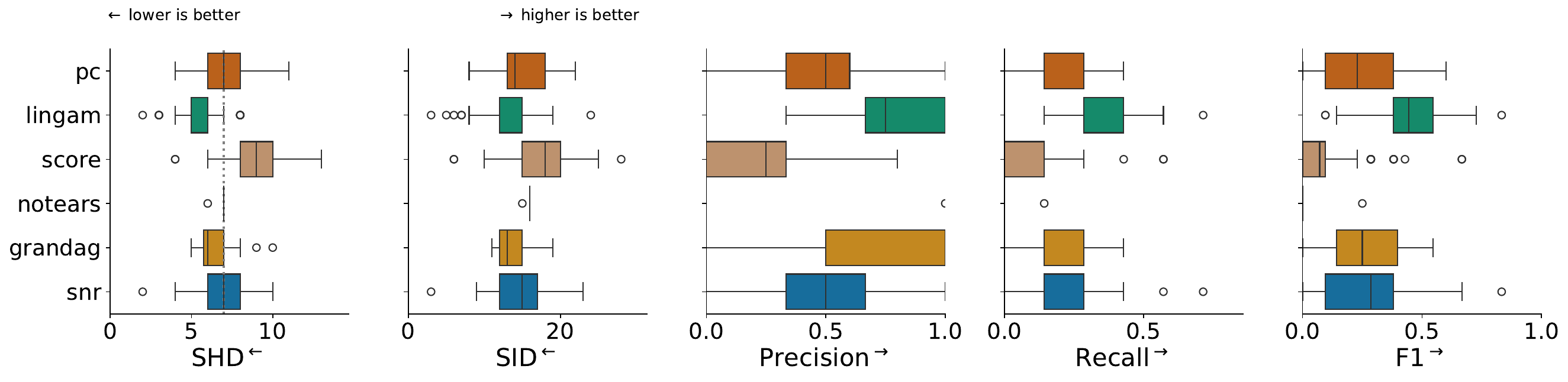}
    \centering
    \caption{Chosen metrics for assessing causal discovery outcomes on Station \(1\) (\(\lvert V_1 \cup V_2\rvert = 6\)) using \(100\) simulation runs on a sample size of \(n = 500\).}
    \label{fig:station1_benchmark}
\end{figure}

\begin{figure}
    \includegraphics[width = \linewidth]{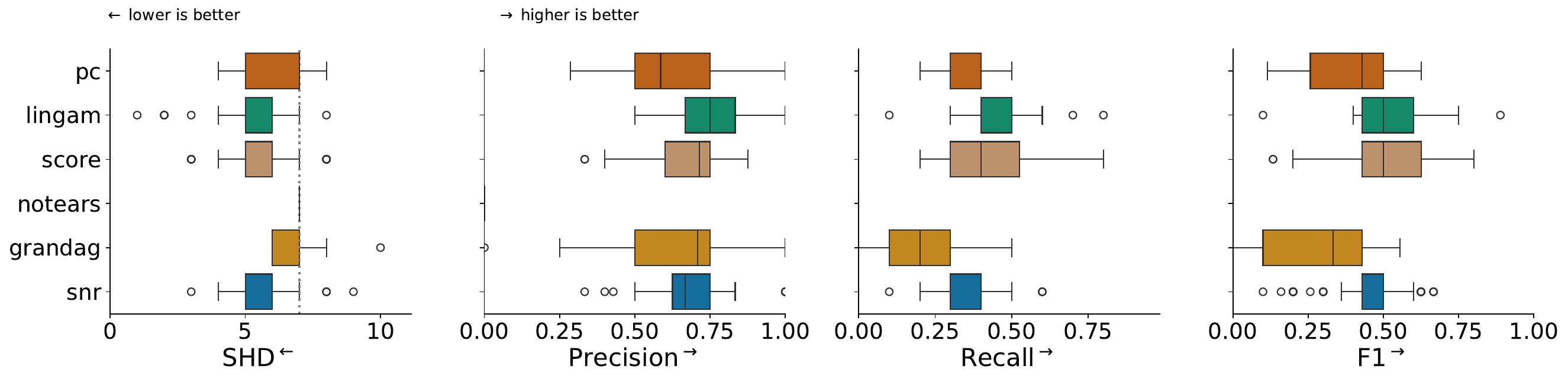}
    \centering
    \caption{Chosen metrics for assessing causal discovery outcomes on Station \(1\) (\(\lvert V_1 \cup V_2\rvert = 6\)) after converting all DAGs to CPDAGs using \(100\) simulation runs on a sample size of \(n = 500\).}
    \label{fig:station1_cpdag_bench}
\end{figure}

We find that results vary substantially across stations. Figure \ref{fig:station1_benchmark} and \ref{fig:station1_cpdag_bench} report results for the first station. DirectLiNGAM performs best while NOTEARS is not able to detect any edges. GraN-DAG results are very unstable throughout all benchmarks. The PC algorithm performs only slightly worse than DirectLiNGAM and SCORE's performance dropped when transformed to the corresponding MEC.

\begin{figure}
    \includegraphics[width = \linewidth]{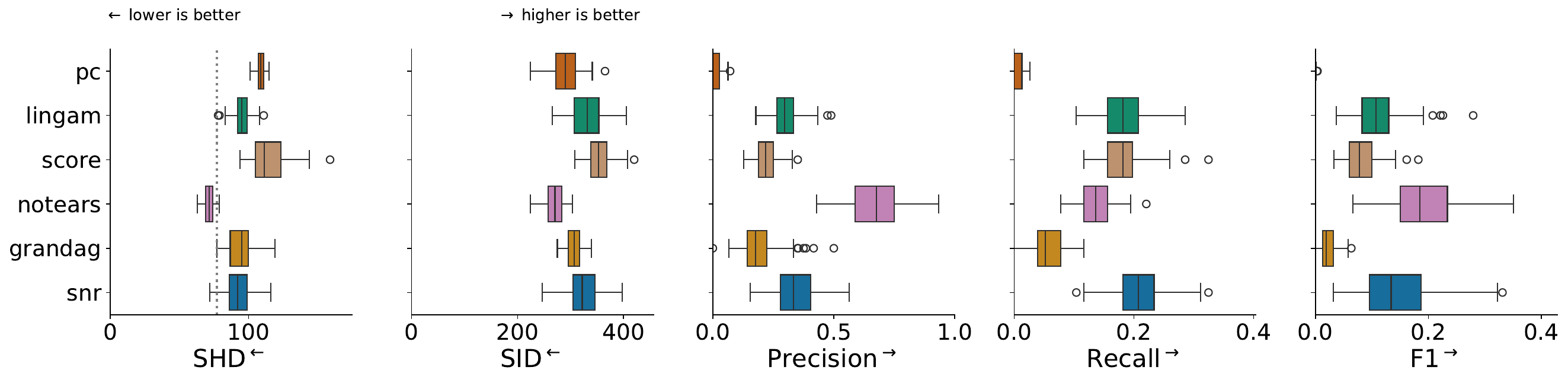}
    \centering
    \caption{Chosen metrics for assessing causal discovery outcomes on Station \(2\) (\(\lvert V_3 \cup V_4 \rvert = 34\)) using \(100\) simulation runs on a sample size of \(n = 500\).}
    \label{fig:station2_benchmark}
\end{figure}

\begin{figure}
    \includegraphics[width = \linewidth]{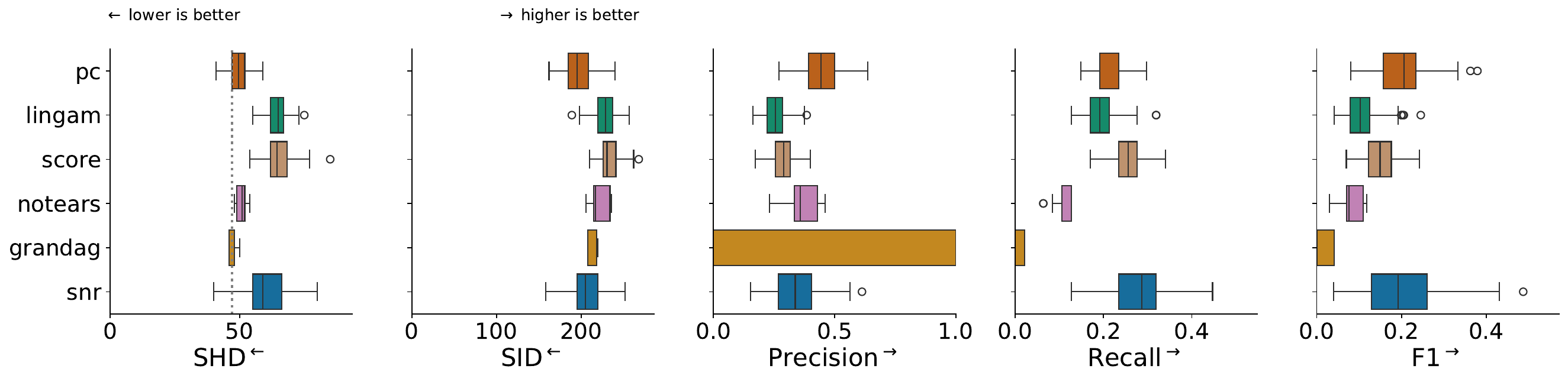}
    \centering
    \caption{Chosen metrics for assessing causal discovery outcomes on Station \(4\) (\(\lvert V_7 \cup V_8 \rvert = 26\)) using \(100\) simulation runs on a sample size of \(n = 500\).}
    \label{fig:station4_benchmark}
\end{figure}

\begin{figure}
    \includegraphics[width = \linewidth]{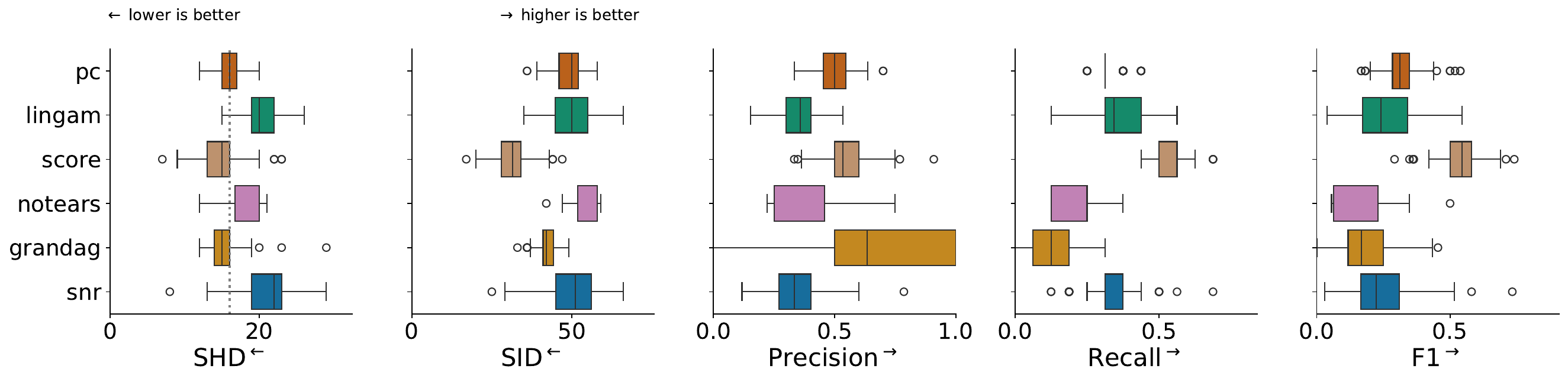}
    \centering
    \caption{Chosen metrics for assessing causal discovery outcomes on Station \(5\) (\(\lvert V_9 \cup V_10 \rvert = 16\)) using \(100\) simulation runs on a sample size of \(n = 500\).}
    \label{fig:station5_benchmark}
\end{figure}

\begin{figure} 
    \includegraphics[width = \linewidth]{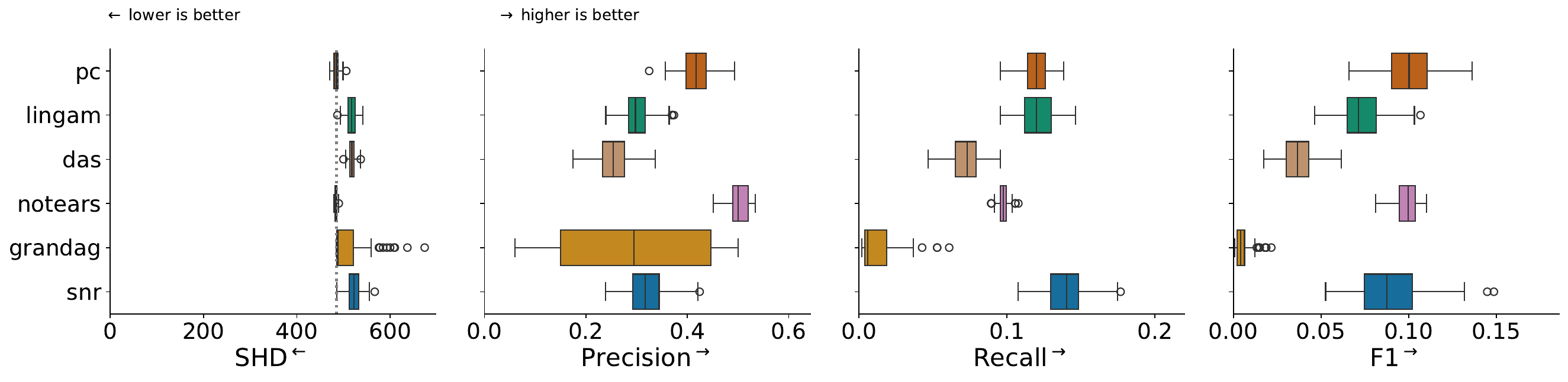}
    \centering
    \caption{Chosen metrics for assessing causal discovery of the full assembly line based on CPDAG comparisons (\(\lvert V \rvert = 98\)) using \(100\) simulation runs on a sample size of \(n = 5000\). The dashed line corresponds to the SHD between the empty graph and ground truth indicating the number of edges in the ground truth CPDAG.}
    \label{fig:full_cpdag_benchmark}
\end{figure}

\begin{figure}
    \includegraphics[width = \linewidth]{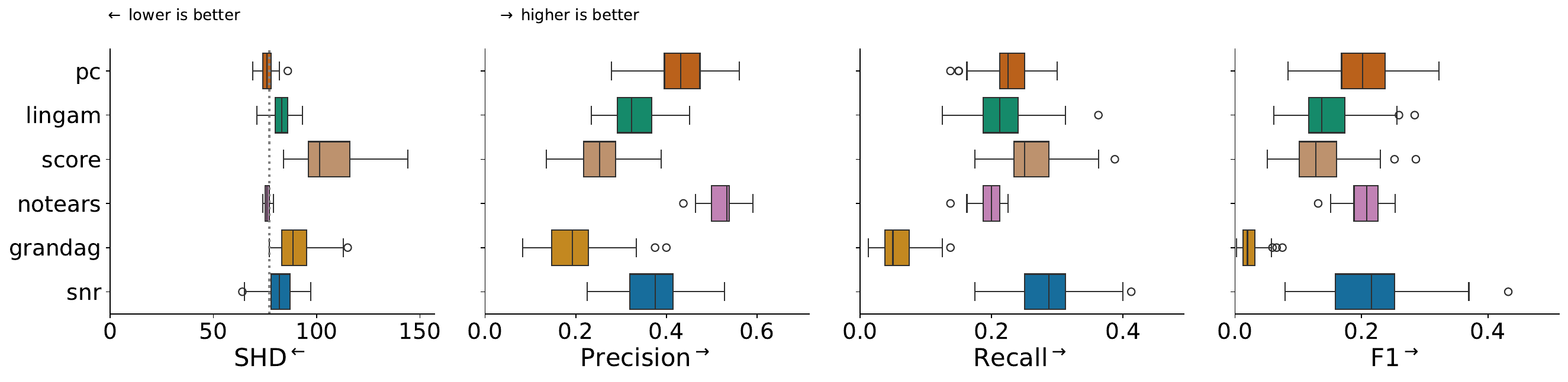}
    \centering
    \caption{Chosen metrics for assessing causal discovery outcomes on Station \(2\) (\(\lvert V_3 \cup V_4 \rvert = 34\)) after converting all DAGs to CPDAGs using \(100\) simulation runs on a sample size of \(n = 500\).}
    \label{fig:station2_cpdag_bench}
\end{figure}

\begin{figure}
    \includegraphics[width = \linewidth]{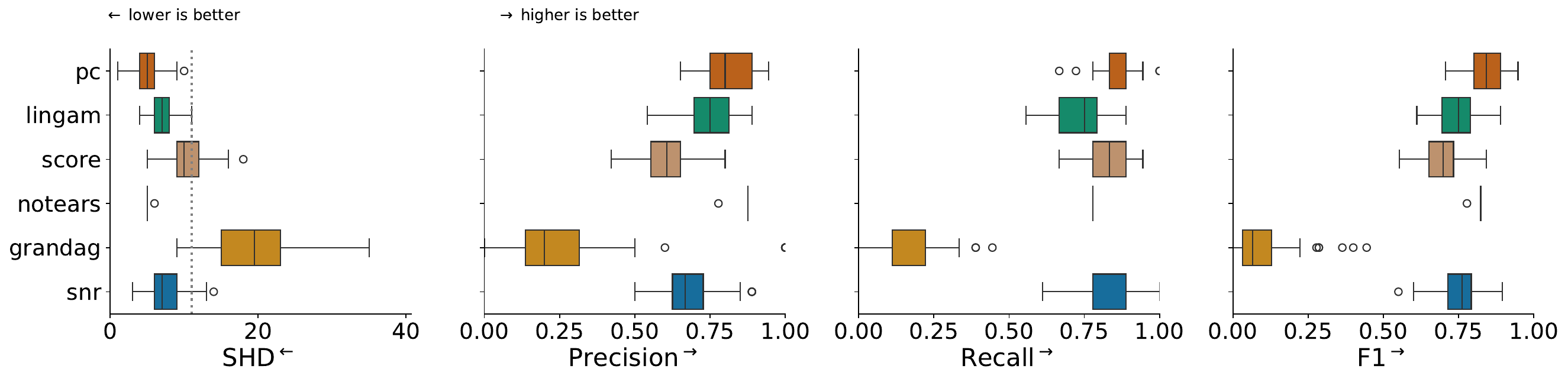}
    \centering
    \caption{Chosen metrics for assessing causal discovery outcomes on Station \(3\) (\(\lvert V_5 \cup V_6 \rvert = 16\)) after converting all DAGs to CPDAGs using \(100\) simulation runs on a sample size of \(n = 500\).}
    \label{fig:station3_cpdag_bench}
\end{figure}

\begin{figure}
    \includegraphics[width = \linewidth]{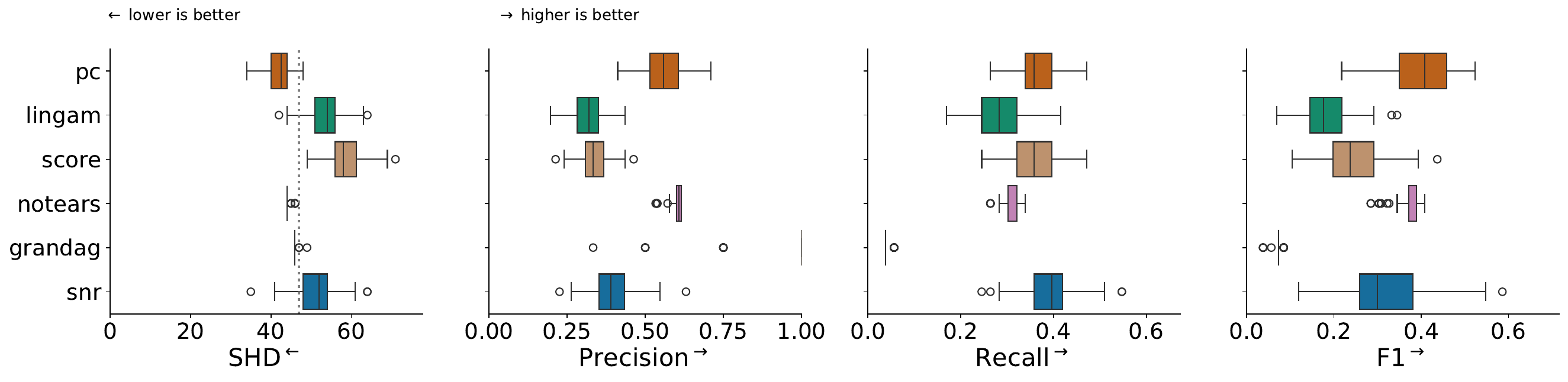}
    \centering
    \caption{Chosen metrics for assessing causal discovery outcomes on Station \(4\) (\(\lvert V_7 \cup V_8 \rvert = 26\)) after converting all DAGs to CPDAGs using \(100\) simulation runs on a sample size of \(n = 500\).}
    \label{fig:station4_cpdag_bench}
\end{figure}

\begin{figure}
    \includegraphics[width = \linewidth]{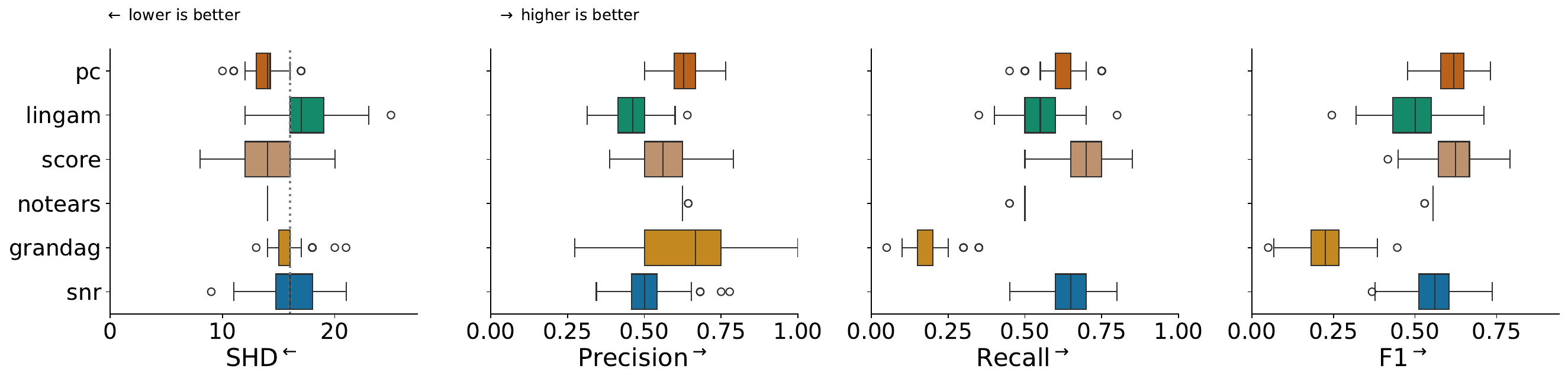}
    \centering
    \caption{Chosen metrics for assessing causal discovery outcomes on Station \(5\) (\(\lvert V_9 \cup V_10 \rvert = 16\)) after converting all DAGs to CPDAGs using \(100\) simulation runs on a sample size of \(n = 500\).}
    \label{fig:station5_cpdag_bench}
\end{figure}

In Figure \ref{fig:station2_benchmark} and \ref{fig:station2_cpdag_bench} Station \(2\) two benchmarks are depicted. In this case, NOTEARS performs best in the context of recovering the true DAG. When converted to CPDAGs, NOTEARS performance drops and PC algorithm performance increases drastically. Figure \ref{fig:station3_cpdag_bench} reports Station \(3\) benchmarks for the CPDAG harmonization strategy. In this case  the PC algorithms outperforms all other routines when compared to the ground truth CPDAG. In Figures \ref{fig:station4_benchmark} and \ref{fig:station4_cpdag_bench} Station \(4\) benchmarks suggest again that the PC algorithm outperforms all other causal discovery algorithms. On station five (Figure \ref{fig:station5_benchmark} and \ref{fig:station5_cpdag_bench}) SCORE overall performs best on DAG level and the PC algorithm draws equal for the CPDAG harmonization strategy. Laslt, Figure \ref{fig:full_cpdag_benchmark} illustrates the full line benchmark results for the CPDAG harmonization strategy. As mentioned in the main text, NOTEARS performs best overall, GraN-DAGs behavior is very unreliable, and the PC algorithm is the runner-up. Yet, none of the procedures is able to perform significantly better than \textit{random regress}.

\clearpage\newpage
\subsection{Hyper and tuning parameters}\label{sec:tuning}

\begin{longtable}{p{.50\textwidth} p{.50\textwidth}}

    \caption{Hyper and tuning parameter choices for routines employed in this work.\label{table:tuning}} \\
    \toprule                                                                                             \\
                 & Parameters                                                                            \\ \midrule
    SpAM         & \texttt{num\_folds} = \(5\)                                                           \\
                 & \texttt{cv\_loss} = MSE                                                               \\
                 & \texttt{lambda\_choice} = 1se                                                         \\
                 & \texttt{num\_bsplines} = \(6\)                                                        \\ [2em]
    DRF          & \texttt{min\_node\_size} = \(15\)                                                     \\
                 & \texttt{num\_trees} = \(2000\)                                                        \\
                 & \texttt{splitting\_rule} = FourierMMD                                                 \\ [2em]
    PC Algorithm & \texttt{indep\_test} : fisherz                                                        \\
                 & \texttt{alpha} = \(0.05\)                                                             \\ [2em]
    NOTEARS      & \texttt{lambda\_1} = $0.1$                                                            \\
                 & \texttt{loss\_type} = $\ell_2$                                                        \\
                 & \texttt{max\_iter} = \(100\)                                                          \\
                 & \texttt{h\_tol}  = \(1\mathrm{e}-8\)                                                  \\
                 & \texttt{rho\_max} = \(1\mathrm{e}+16\)                                                \\
                 & \texttt{w\_threshold} = \(0.3\)                                                       \\ [2em]
    GraN-DAG     & \texttt{input\_dim} = \(V\)                                                           \\
                 & \texttt{hidden\_num} = $2$                                                            \\
                 & \texttt{hidden\_dim} = $10$                                                           \\
                 & \texttt{batch\_size} =$64$                                                            \\
                 & \texttt{lr} = $0.001$                                                                 \\
                 & \texttt{iterations} = $10000$                                                         \\
                 & \texttt{model\_name} = NonLinGaussANM                                                 \\
                 & \texttt{nonlinear} = leaky-relu                                                       \\
                 & \texttt{optimizer} = rmsprop                                                          \\
                 & \texttt{h\_threshold} = $1\mathrm{e}-8$                                               \\
                 & \texttt{lambda\_init} = $0.0$                                                         \\
                 & \texttt{mu\_init} = $0.001$                                                           \\
                 & \texttt{omega\_lambda} = $0.0001$                                                     \\
                 & \texttt{omega\_mu} = $0.9$                                                            \\
                 & \texttt{stop\_crit\_win} = $100$                                                      \\
                 & \texttt{edge\_clamp\_range} = $0.0001$                                                \\ [2em]
    SCORE, DAS   & \texttt{eta\_G} = $0.001$                                                             \\
                 & \texttt{eta\_H} = $0.001$                                                             \\
                 & \texttt{alpha} = $0.05$                                                               \\
                 & \texttt{n\_splines} = $10$                                                            \\
                 & \texttt{splines\_degree} = $3$                                                        \\
    \bottomrule
\end{longtable}

Table \ref{table:tuning} reports all hyperparameters and tuning parameter choices for benchmark results as well as for SpAM and DRF fitting. All benchmark algorithms were taken off-the-shelf to avoid providing any of the procedures with an unfair advantage.

\end{document}